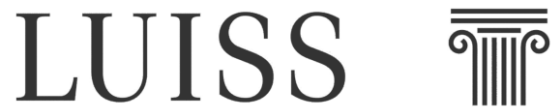

*Department of* Business and Management

*Bachelor's Degree:* Management and Computer Science

*Course:* Artificial Intelligence and Machine Learning

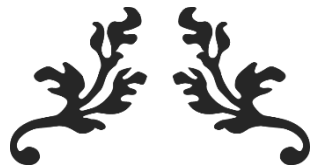

# MACHINE LEARNING MEETS MENTAL TRAINING

A Proof of Concept Applied to Memory Sports

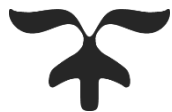

*Supervisor:*

Professor Giuseppe Francesco Italiano

*Candidate:*

Emanuele Regnani

ID No.: 251961

*Academic Year:* 2022 – 2023

# TABLE OF CONTENTS

# INTRODUCTION

*"Mens sana in corpore sano"* (Juvenal, 100-127 AD)

Mental training has long been part of human culture, appearing in several different forms ranging from meditation to particular games or cognitive exercises aimed at various purposes. The past decades, however, have seen it losing its cardinal role in the well-roundedness of an individual and becoming more of a side hustle, confined to particular hobbies or to specific techniques needed for mental-health purposes. By contrast, recent years have seen an exponential investment in and development of artificial intelligence and machine learning technologies, which seem to be successfully tackling increasingly difficult tasks and problems.

This work, then, aims to combine the two fields together by presenting a practical implementation of machine learning to the particular form of mental training that is the art of memory, taken in its competitive version called "Memory Sports". Such a fusion, on the one hand, strives to raise awareness about both realms, while on the other it seeks to encourage research in this mixed field as a way to, ultimately, drive forward the development of this seemingly underestimated sport.

After first introducing the topic of mental training and its particular branch of Memory Sports, in the first chapter, the machine learning involved in the project is explained in the second chapter. The third chapter, then, presents two practical implementations of machine learning in Memory Sports, the results of which are discussed in the final chapter, together with several potential directions for future research.

Ultimately, as well as stimulating interest and inspiration regarding the two fields involved in this research and exploring their points of contact, the aim here is also to investigate potential developments of human-machine collaborations, which are likely to be the focus of the next advances in technology and society overall. Starting to think in this view can help better prepare for the abrupt changes that might come and even be part of them, so as to drive their aim and scope toward a more responsible, and thus better, outcome.

CHAPTER ONE

# MENTAL TRAINING AND MEMORY SPORTS

## 1. Mental Training and the Art of Memory

### 1.1. Mental Training

> *"Glasses are like a mask we wear, a way of hiding our true selves from the world. They create an illusion of clarity and precision, but in reality, they only distort our vision. They are a kind of deception, a way of pretending to be something we're not."* (Svevo, 1923).

Like all life forms, humans once used to struggle to survive in their environment, needing to train and sharpen a wide range of skills, including hunting and collaboration. Modernity has gradually eased things, by providing virtually infinite food supplies and sources of entertainment. Yet, we were not made for this at all: the human body and mind evolved in a completely different environment requiring both physical and mental abilities that we seem to be no longer required to possess. On the other hand, for survival purposes, we developed a pressing tendency to save on the usage of resources whenever possible. The resulting internal conflict between this constant attempt to save resources and the intrinsic need of our bodies to move has led to a world in which, by reducing physical movement to the bare necessary, we are actually harming ourselves, so that more people die from obesity than from hunger[1]. A similar, and perhaps even more pronounced, phenomenon happens with the mind, where attention disorders and dementia are skyrocketing because we tend to delegate increasingly more to our technological devices, resulting in us often no longer making meaningful use of our cognitive faculties.

---

[1] *https://edition.cnn.com/2012/12/13/health/global-burden-report/index.html*

The situation has not gone completely unnoticed, fortunately; hence, more and more people are now making physical training a part of their daily lives, or it is at least becoming common sense that one should train their body. The same is not true for mental training, however; because little understanding of the brain's need for it is there among non-experts, common sense seems to ignore the existence of the field entirely. Moreover, unlike physical sports, which have long been a popular form of entertainment and of challenging and testing humanity's limits and capabilities, mental sports are often completely neglected or considered extremely niche or even awkward, with only a few such as chess and the Rubik's cube receiving any attention at all. Finally, yet most crucially, while today's world might not be particularly motor-skills oriented, it is instead more than ever rewarding mental abilities, be it through contests, talent shows, or simply the academics world and the job market. Hence, it appears as if we are moving in the opposite direction with respect to what we then require of ourselves.

Mental training has a long history and tradition, taking various forms and focusing on many different aspects; its main instance is represented by meditation, but also the martial arts play a vital role in the scene, particularly by connecting mental and physical training together. Other examples, while initially just a game, can and often do evolve into particularly articulate mental exercises and routines; these include the aforementioned chess but also Go, which has a particularly renowned tradition in Asia. Many have to do with solving particular puzzles, such as the Rubik's cube, mental math, or even Sudoku. The one which will be central to this thesis is the art of memory, specifically its competitive version that is Memory Sports. Before diving deeper, however, let us stress slightly more the importance of training one's mind in general.

The mind is the filter through which all our reality is perceived; its state and our ability to control it, therefore, directly influence our perception of the world and of our life itself. This is why mindset is believed to be so crucial in several realms and this is why loss of focus or inability to manage emotions can prove so detrimental on many occasions. A trained mind, then, can aid tremendously in concentrating in the task at hand and reaching the deep-work flow states necessary to deliver valuable results; it can help in channeling the tension felt upon giving a public speech into the energy needed to engage one's audience; it can even just simply be the difference between seeing the glass half empty and half full in challenging emotional situations. Be it for the improved performance and decision-making skills it brings or for the emotional control it allows for, mental training is ultimately a matter of experiencing the full spectrum of potential our minds are capable of and, armed with that knowledge and skill, living a more intentional and hence happier life.

## 1.2. The Art of Memory

**Writing and the Internet**

> *"If men learn this, it will implant forgetfulness in their souls. They will cease to exercise memory because they rely on that which is written, calling things to remembrance no longer from within themselves, but by means of external marks."* (Plato, 370 BCE)

When humans first used writing, it was likely due to a memory overload issue: as explained by Y.N. Harari in his bestseller "Sapiens", large systems of cooperation involving not few but thousands or even millions of humans require the handling and storage of huge amounts of information, much more than any single human brain can contain and process. Apart from the limited capacity and lifespan of human brains, however, their most crucial limiting factor was their selectiveness in which type of information to store and process. In fact, while evolutionary pressures have adapted the human brain to store immense quantities of botanical, zoological, topographical and social information, they never accustomed it to handle the completely new type of information made vital by the emergence of particularly complex societies in the wake of the Agricultural Revolution: numbers.

Thus, by addressing this exact issue, writing appears to have been an overall beneficial invention, constituting a deeply impactful turn in the history of mankind. As with any tool, however, its potential can broaden both in positive and negative directions. As previously explained, writing was invented to cope with functions the human brain was missing; in order to manage the resulting complexity of systems developed based on this form of encoding information, however, the situation tended to overturn: whilst humans had initially invented writing to accommodate their own needs, it is as if they subsequently had to adapt themselves to its own way of being, as they were the most flexible out of the two. Those who operated systems of this kind, in fact, had to start thinking in a non-human fashion, similarly to filing cabinets instead; free association and holistic thought have given way to compartmentalization and bureaucracy, as condensed by Harari. As a result of this, according to the author, humans already are and increasingly will be dependent on and led by numbers and computers, who will eventually overrule them completely: *"The process that began in the Euphrates valley 5,000 years ago, when Sumerian geeks outsourced data-processing from the human brain to a clay tablet, would culminate in Silicon Valley with the victory of the tablet. Humans might still be around, but they could no longer make sense of the world. The new ruler of the world would be a long line of zeros and ones."*. Whilst this latter might appear as a slightly extreme and conspiratorial view, it does effectively warn us of the fully plausible consequences such a powerful tool holds in potential.

These warnings are particularly relevant in today's age of information: every quirk, notion, or detail seems to be readily accessible through online encyclopaedias or other sources. Yet, as helpful as these instruments can be, they do not represent knowledge or wisdom by themselves: *"what you have discovered is a recipe not for memory, but for reminder. And it is no true wisdom that you offer your disciples, but only the semblance of wisdom, for by telling them of many things without teaching them you will make them seem to know much while for the most part they know nothing. And as men filled not with wisdom but with the conceit of wisdom they will be a burden to their fellows."* (Plato, 370 BCE). Knowledge, and most importantly wisdom, comes instead from the deep interiorization of concepts, from their "fermenting" into one's mind resulting in random and spontaneous connections between them. Knowledge also results from testing and applying such ideas that come to mind through this process, as this way they are deepened and developed even further, triggering the virtuous cycle between direct experience and reflection that ultimately leads to profound wisdom.

This "lively" characteristic data acquires once in a human mind is ultimately what opposes it from it lying on a fixed external device: *"you know, Phaedrus, that is the strange thing about writing, which makes it truly correspond to painting. The painter's products stand before us as though they were alive. But if you question them, they maintain a most majestic silence. It is the same with written words. They seem to talk to you as though they were intelligent, but if you ask them anything about what they say from a desire to be instructed they go on telling just the same thing forever."* (Plato, 370 BCE). Memories are these very pieces of data we work with, they are the concepts that make up our understanding, constituting both the basis and the result of our thinking and knowledge: *"memory is the residue of thought."* (Brown, et al., 2014). Thus, the way we register and encode such ideas in our brains directly determines our future understanding of the larger concepts they make up, as well as to what extent we are then able to use them to think and reason. A weak encoding results in such notions quickly evaporating or not being well connected to previous knowledge, things without which there can be no learning: *"'t is not knowledge, The having heard without retaining it."* (Alighieri, 1320)

**The Art of Memory**

In Greek mythology, the goddess of memory Mnemosyne was the mother of the nine muses, representing literature, science, and the arts, and often invoked for inspiration by artists. This was not by accident: in fact, also the ancients believed that *"memory is the mother of all wisdom."* (Aeschylus, 424 BCE). As a result, the art of memory was considered a vital tool for education and public speaking by allowing one to remember large amounts of information, such as speeches, poems, and historical events. Cicero, for instance, believed the ability to remember and recall information to be the foundation for all intellectual and creative pursuits, as well as being essential for understanding and interpreting the world around us, and for communicating our ideas and thoughts to others. He also saw the art of memory as the foundation for other forms of knowledge and creativity, such as philosophy, literature, and the visual arts. Others on this line were the philosopher Aristotle, writing that memory is the *"scribe of the soul"*, or the medieval scholar Thomas Aquinas, who called it the *"storehouse of the intellect."*. It was even used during the Middle Ages to remember entire books, as censorship rules prevented them from being openly spread around, or in various religious contexts to recall the respective sacred books from memory, or in some indigenous cultures to aid in the preservation and transmission of oral histories and cultural knowledge.

Nowadays, the relevance of the art of memory for such feats of intellect has largely declined as a result of technology, yet it continues to serve purposes such as keeping mental agendas, retaining names and faces of newly-met people, or remembering particular details in any circumstance; most importantly, however, it plays a significant role in examinations and academics in general, in quicker and deeper learning, and in memorizing work-related information or directives. These last points are of particular importance in an age of *"increasing complexity and change, where the skills that made us successful yesterday may not be enough to keep us ahead tomorrow. In this world, the ability to learn quickly and effectively is not just an advantage - it's a necessity."* (Young, 2019). With new technologies constantly being deployed and entire industries being revolutionized all the time, learning itself, rather than learning a particular skill, has truly become today's superpower. People in education today will probably perform jobs that do not even exist yet; thus, adaptation is perhaps the most needed skill of the century.

Memory training does not only aid with retention and recall, however. Due to its particular nature of generating memorable and innovative images and stories (which will be partly expanded on later), it spurs creativity and neuroplasticity, as well as a general ability to connect different ideas and concepts together, which we have seen is key to learning and to intelligence in general. This is also one of the most crucial realms where AIs do and will struggle to replace humans in the workforce, according to Richard Baldwin's research in "The Globotics Upheaval". Overall, it makes the brain work better as a whole, making one sharper and brighter. Being a mental sport in itself, then, it also develops in the practitioner a range of traits typical of cognitive training in general, the most important of which is focus. By requiring one to set aside everything during the training or recall, to channel all their mental energies toward a single task, and thus to momentarily control one's thoughts and emotions, a training of this kind ultimately develops the control of one's attention, which together with the ability to *"quickly master hard things"* referred to above makes what are considered by the author Cal Newport to be the *"two core abilities for thriving in the new economy"*. In particular, he describes this skill of dominating one's focus as *"the ability to produce at an elite level, in terms of both quality and speed."* resulting from *"a state of distraction-free concentration when your brain works at its maximum potential"*, a state of flow that he calls "deep work". Attention, the author continues, has also been demonstrated to be the lens through which we construct our worldview, so much that *"what we choose to focus on and what we choose to ignore—plays in defining the quality of our life."*. Finally, the author also highlights the value of the combination of these two qualities of innovation and attention, stating that *"[great creative minds] think like artists but work like accountants."*

Some other strengths of engaging in mental or memory training are: its proven ability to fight and slow down aging, particularly with regards to dementia and similar conditions; the healthy competition and personal growth it encourages, especially in the case of its competitive side of Memory Sports; the network and environment of stimulating, brilliant, and insightful people it gives access to, both online and in person. Most importantly in all of this, however, is that memory training ultimately teaches a way to think (that is, through images and associations) that goes beyond its applications to memory and instead unlocks the true potential of the mind by strengthening understanding, learning, and reasoning, thanks to the processes mentioned above.

# 2. Memory Sports and Technology

## 2.1. Memory Sports

Apart from its application in academic or daily-life contexts, the art of memory also expresses itself in a competitive form, a mental discipline referred to as "Memory Sports". Due to its relatively new and unsaturated nature, it is characterized by a lively and vibrant community, eager to explore and experiment with new techniques and to share their findings on the many virtual platforms available. Despite the millennial history of the art of memory, in fact, its application to truly competitive environments is rather recent; hence, novel or even revolutionary methods keep being formulated, from time to time. What ultimately brings this community together is the desire of people to challenge themselves and to, collectively, really test the limits of the human brain through the use of the internal support devices that are memory techniques.

The sport manifests itself in several disciplines, the most common of which being: Numbers, Cards, Names and Faces, Words, Images. Other popular ones are: Binary Digits, Dates and Events, Spoken Numbers (where numbers are dictated out loud for competitors to memorize on the spot). Finally, regional competitions tend to also present some peculiar disciplines of their own, namely: "Exam Cramming"[2], a challenging mix of fictional geography, history, foreign languages, identification, and alternative science; "Three Strikes You're Out"[3], which involves memorizing diverse facts about strangers; "It's Been a Long Time"[4], involving long-term recall of previously provided data; Poetry. All these diverse disciplines also present different formats and variations, having to do with factors as their length, or mode of recalling the data, the details of which will not be dealt with here, however.

The most common and widespread types of competitions, each with their unique combination of the available disciplines, are the following two: the "classical format", consisting of the main disciplines mentioned above arranged in a decathlon lasting one to three days; a head-to-head format including only the first group of disciplines cited earlier. The former is typical of the "International Association of Memory (IAM)"[5], the main entity in the field which also organizes the yearly World Memory Championships, whilst the latter has been popularized by the "Memory League"[6] website and platform. The specifics of each format, as well as the types of tournaments and seasons that take place, are out of the scope of this text; however, further details can be found in an Italian article[7] by the IAM World Memory Champion Andrea Muzii or on the "Art of Memory" forum[8].

These same resources also prove valuable when it comes to the specific techniques used by memory athletes to perform such feats of memory. As anticipated above, in fact, such competitors are able to "hack" their natural memory by simply exploiting proven principles regarding its functioning, which are then trained so as to become second nature. Because of the universality of these principles, then, anyone can learn to apply such techniques to a surprising extent within just a few weeks or months of deliberate training. Due to the scope limitations of this text, however, only two of the main ones will be outlined here: the memory palace and systems.

---

[2] *https://canadianmindsports.com/memory-sports/*
[3] *https://www.usamemorychampionship.com/*
[4] *https://www.usamemorychampionship.com/*
[5] *https://www.iam-memory.org/*
[6] *https://memoryleague.com/#!/home*
[7] *https://blog.andreamuzii.it/diventare-un-memory-athlete/*
[8] *https://forum.artofmemory.com/*

**Memory Techniques**

The basic principle behind memory techniques is to exploit the types of memory that we humans are already proficient at to compensate for those we are not. For instance, the main way data is memorized by practitioners of these methods is by literally making it more memorable than it would be by itself, exaggerating certain aspects of it or even completely transforming it into a more memorable form. Such a practice, of course, requires intense training and empirical experimentation simply to truly internalize all the diverse nuances and expressions of this concept of "memorability".

The memory palace, or method of loci, is a powerful technique making use of internal mnemonic devices called, indeed, "palaces" or "journeys" to store specific pieces of information in different locations within those spaces, which are already familiar to the user beforehand. When mentally moving through a journey, then, they can recall the information placed there in its specific order. Such a technique works by exploiting the power of our spatial memory, which automatically records most of the paths we go through and the structures we visit, together with the specific position of elements such as pieces of furniture or other objects. By combining the intrinsic memory one has of a place they have previously visited with novel information they want to remember, therefore, they are able to essentially use the former as a hook on which to bind the latter, which can then be seamlessly recalled by mentally going through the already present memory of the location used as a palace.

Another technique to mention for the purposes of this work is one rather advanced in terms of its use, yet not so complex in its essence. In order to speed up the memorization of particularly "unremarkable" pieces of data such as numbers and playing cards, in fact, memory athletes tend to build systems of prepared images to replace such data with. Since they would have to transform such information into more memorable forms anyway, the idea is to learn a conversion system by heart so that each number or playing card can be instantaneously memorized by them in its alternative more evocative form. Thus, for instance, athletes end up having ready-to-use mental images for each two-digit or even three-digit number, and for each poker card or pair of poker cards, which evaluates to hundreds if not thousands of these entries.

Having now introduced the world of Memory Sports and of the techniques used by these athletes, let us explore the relationship this sport has with technology.

## 2.2. State of the Art: Current Applications of Technology to Memory Sports

Memory Sports are inherently bound to technology, on which they rely for most of their functioning. In fact, one could say technology has perhaps been the major driving force enabling their development at all.

First and foremost, in fact, Memory Sports are a global community of enthusiasts united by social media platforms and forums; hence, the very essence of the sport (that is, its athletes and fans) would have a hard time connecting to each other otherwise. Because of its relatively small size, in fact, they have rarely managed to simply survive as a local phenomenon, so that, for example, most competitions rely on foreign competitors to even reach the necessary numbers in the first place.

This same "supranational" characteristic of the field enabled by technology is again reliant on IT to sustain itself at all. As much as in-presence events are an integral part of the sport, in fact, complementing them with online ones has proved necessary to fuel it, as physical distances, visas, and resources can truly be a challenge for many competitors in several occasions. This was particularly evident during the COVID-19 pandemic, through which the sport was only able to survive thanks to the rising of the Memory League platform, which allowed both for the athletes themselves to compete virtually, as well as for fans and enthusiasts to follow the events remotely.

Last but not least, tech is crucial in generating data for the athletes to memorize in the first place: with the possibility for one to obtain limitless amounts of random digits, playing cards, or any other data, neatly organized in a grid or in whatever other fashion needed, memory training and competitions have never been this favored. Although some data types such as names or images require a pre-populated database, software now allows to randomly shuffle such records to generate ever novel sequences to memorize, organizing them into the format needed for the training or the competition, and complementing all of this with a program able to keep track of time and correct the athlete's mistakes on the spot. For most competitions and training sessions, long are gone the days where disciplines were printed on paper and arbiters had to manually check if each piece of data recalled by the athlete was exactly matching the original sequence. Additionally, a training software facilitating training and competing in this way greatly stimulates both athletes to train and prepare for competitions, as well as organizers to host competitions themselves, overall profoundly benefitting the sport and its development.

**State of the Art**

Regarding the community side of things, the previously mentioned Art of Memory forum[9] is its heart: here, matches, seasons, and slams are organized, along with ideas and content being shared and discussed all the time. Facebook also acts as an important medium, as the World Memory Championships[10] and the International Association of Memory (IAM)[11] groups are active there, promoting events and raising issues to be discussed with the community. Then, a variety of messaging apps are of course used by the athletes to reach out to one another, in addition to the forum's messaging functionality. Finally, the Twitch[12] streaming platform is used to broadcast head-to-head online matches and tournaments but is also increasingly attempting to do so for in-presence competitions; with graphics and commentators adding to the show, viewers gather together to follow events and witness the athletes challenging themselves and each other.

---

[9] *https://forum.artofmemory.com/*
[10] *https://www.facebook.com/groups/123077091106550/*
[11] *https://www.facebook.com/groups/1753538928199295/*

Whilst usually organized and promoted on the platforms just mentioned, competitions are actually played on the IAM Competition Software[13] (for classical-format ones) and on the Memory League[14] (for head-to-head ones) website. For some particular formats or events, other platforms might be used, but these are the most common ones.

These software programs are also the major ones used for training, as they offer the same disciplines tested in competitions in a suitable training environment. For some particular disciplines or to be exposed to a more varied database, other tools are also considered, namely Standard Memory[15] and MemoCamp[16] among others. Finally, mobile apps like Memory Ladder[17] for Android can be ideal to train without a personal computer.

For one's memory palaces, spreadsheet programs as Excel are mostly used, yet some applications including the Art of Memory software[18] and MemoryOS[19] can be of help. For systems, instead, while one could potentially write them down anywhere, programs akin to Anki[20] prove essential to their interiorization, as they allow to learn them thoroughly and efficiently through the well-documented spaced repetition[21] protocol.

As just shown, the sport lives in a deep symbiosis with computing. In all of this, however, it appears as if the particular technology that is machine learning is yet to enter the scene; the only instance is perhaps represented by the announced plans of the abovementioned MemoryOS app to implement some of the latest AI tools into their software. A rather large section regarding the potential future applications of artificial intelligence to the field concludes this research, then, precisely for the purpose of addressing this shortcoming. For now, however, having explored the current situation, let us finally move onto the research question guiding this project.

---

[12] *https://www.twitch.tv/memorysportstv*
[13] *https://www.iamwmc.com/competition/training.html*
[14] *https://memoryleague.com/#!/home*
[15] *https://www.standard-memory.com/*
[16] *https://memocamp.com/en*
[17] *https://play.google.com/store/apps/details?id=com.mastersofmemory.memoryladder&hl=en&gl=US&pli=1*
[18] *https://artofmemory.com/app/*
[19] *https://memoryos.com/mind-palace-virtual-implant*
[20] *https://apps.ankiweb.net/*
[21] *https://en.wikipedia.org/wiki/Spaced_repetition*

# 3. Research Question

## 3.1. The Scoring System in Memory Sports

As stated above, there are two main competition formats in vogue, which from now on will be referred to as the "IAM" format and the "Memory League" one, for brevity. Apart from them differing in terms of the combination of disciplines they comprise and in the way opponents face each other (all-against-all vs head-to-head), their major difference for the purposes of this text is in their scoring system. Whilst the Memory League format simply detracts a point for each mistake, resulting in a final score equal to the total number of correctly memorized pieces of data, the IAM one is much harsher, highly penalizing the athlete for each mistake.

Another difference has to be pointed out: whilst the IAM format is usually characterized by fixed memorization times, during which a variable amount of data can be memorized based on the athlete's pace, the Memory League one rather has fixed quantities of data to memorize, so that one can also finish earlier than the allotted one minute of time. As a result, whilst IAM competitions are mainly played on quantity, Memory League ones are primarily played on time. These intricacies each give rise to particular equilibria the athletes have to aim at while competing.

## 3.2. The concept of "perfect balance"

As anticipated, the IAM scoring system is particularly penalizing for the mistakes made; at the same time, however, one has to aim for the highest possible quantity of data memorized, so as to beat the other competitors' scores. This duality of conflicting objectives leads to the athlete seeking a particular balance when competing: memorizing as much data as possible, whilst also being as accurate as possible in its encoding. This will here be referred to as the "perfect balance" and will be the core of this research.

A similar situation plays out in the Memory League format: because the fixed quantity of data is usually memorized by advanced athletes in much less than the given one minute of memorization time, the race moves on the temporal axis, with each competitor attempting to take as little time as possible to memorize that same fixed quantity of data. Thus, the question becomes a matter of finding the lowest time at which one can memorize that given quantity of data, while making no mistake. Independently of the time taken, in fact, the quantity is evaluated first when assessing the winner of a head-to-head match; hence, it must remain a priority to achieve the full score. To rephrase, then, while keeping the memorized quantity at its highest possible, one has to find the lowest time they can achieve such feat in. Naming the above one the "IAM perfect balance", this one will then be referred to as the "Memory League perfect balance" and will be the focus of the second task of this research.

Since these are both essentially optimization problems, they represent ideal scenarios to be, at least partially, solved through machine learning; therefore, they make optimal candidates for a research of this kind. Before further explaining the relevance and choice of this exact matter, let us better define the technical details of such pursuit.

## 3.3. Exact Scope of the Machine Learning Program

A small remark to be made here is that, in the act of memorizing, these two different equilibria essentially feel the same, as in both cases one is attempting to go as fast as possible whilst still maintaining the highest possible level of accuracy. Also in the case of the IAM format, in fact, memorizing as much data as possible still equates to going as fast as possible. However, from a technical analysis point of view, these two scenarios do present themselves differently, especially due to the different data formats they provide; they also follow different trends and present some distinct intrinsic characteristics. Therefore, they will be treated separately, despite both essentially stemming from a single, almost "philosophical", matter of balancing speed and accuracy.

Having said that, the research will thus consist of two separate machine learning programs: one for the IAM format and one for the Memory League one.

The first will be a more "universal" software that, given the results of a previous "imperfect" memorization attempt, will yield the memorization quantity to aim at so as to achieve a perfect score (one with no mistakes at all, which is the objective explained earlier), no matter the particular athlete in question. It will do so by attempting, given several training examples, to capture the true nature of the relationship between these three factors: an attempt's quantity of correctly memorized data, its score (resulting from the quantity memorized and the penalization applied to it for the errors made), and the subsequent ideal quantity one should aim at so as to make no mistake. For instance, one might attempt to memorize 200 digits in the given time but only manage to correctly recall 196, most likely resulting in a score of $120^{22}$; the ideal quantity to aim at next time, so as to be reasonably sure to achieve a perfect score, would therefore probably be around 180. The exact way this will be achieved is to be explained in the third chapter, however a key component of it will be a manual selection of the data and of the correlation between attempts and "perfect scores".

The second machine learning program, instead, will be more individual, as it will work with the data of one single athlete at a time. Given their past results in matches on the Memory League website, it will attempt to extract the underlying curve representing their ability; that is, it will try to predict the most probable score for them to obtain for each amount of time taken for the memorization. To illustrate, such curve would be able to tell that one is to get a perfect score as long as they take at least 30 seconds for that particular memorization, whilst they would probably make one mistake at 28 seconds, two mistakes at 27, and so on, in a likely exponential fashion. What is more, once such curves have been obtained for multiple athletes, they could be compared so as to obtain forecasts or to spot particular trends.

A final remark regards the Memory Sports disciplines to be taken into consideration: for the purposes of this research, only the most popular and, according to many, important (in terms of training) will be analyzed; that is, the IAM "5-minute Numbers" or "Speed Numbers" discipline for the first program and the "Numbers" one for the Memory League program. However, this work serves as a proof of concept of what is possible and, hence, it intends to pave the way for future expansions and developments of itself, a matter which will be deepened in the conclusions.

---

[22] the IAM scoring follows slightly peculiar rules which attempt to minimize the impact of consecutive mistakes, as opposed to ones spread apart; hence, the exact score one obtains given the same ratio of correct data to total data actually depends on the distribution of these mistakes in space. The details of these rules will not be explained in this context, but can be further deepened at *https://www.iam-memory.org/wp-content/uploads/2019/11/IAM_Rulebook_Nov2019.pdf*.

## 3.4. Choice and Relevance of the Research Question

As pointed out in the state-of-the-art part of this chapter, this sector brims with potential applications of machine learning, which will be further discussed in the concluding chapter of this research. In order to narrow down the focus, however, and find a specific topic to work on, the challenge of finding the speed to aim at in competitions was chosen here as the ideal combination of feasibility and actionability, usefulness, and quantifiability of both the input data and the results. Regarding its relevance, in fact, it should be stressed that this is a crucial matter in the field, as one's performance in competitions ultimately depends on it; hence, any help in assessing such ideal speed to aim at beforehand can be vital, as it allows the athlete to maximize the results they can obtain, given their current skill level. A tool like this one, therefore, can make the difference in achieving a podium position or not in the scoreboard, or even in winning a head-to-head match at all.

# CHAPTER TWO

# MACHINE LEARNING INVOLVED

*"Machine Learning is the science (and art) of programming computers so they can learn from data"* (Géron, 2019), so they can learn without being explicitly programmed themselves. It works by using some examples as "training data" to teach the system a particular relation between them or the ideal way to perform a particular task, after which the system's performance is evaluated on a "testing set" of data of the same kind, according to the most appropriate performance measure for the task at hand. The peculiarity of such a system, as compared to one resulting from traditional programming techniques, is its ability to automatically learn, to spontaneously fit itself to the data provided and understand the underlying relation connecting the data points.

As a result of its nature, machine learning usually results in shorter, easier to maintain, and most likely more accurate programs. However, its main advantages lie in its specific applications: it is, in fact, of particular help in problems that either are too complex for traditional approaches or have no known algorithm, such as speech recognition; it is also optimal for changing environments, as such a system can adapt to new data. Finally, machine learning *"can help humans learn: ML algorithms can be inspected to see what they have learned (although for some algorithms this can be tricky). […] Sometimes this will reveal unsuspected correlations or new trends, and thereby lead to a better understanding of the problem."* (Géron, 2019). No wonder the at-the-time World Chess Champion Garry Kasparov, first beaten by an AI technology in 1997, later expressed the idea of the ultimate player being a human-machine cyborg, optimally combining the abilities of both[23].

A typical machine learning project starts with gathering and studying the data, often involving manipulation and calibration of these latter. It then continues with the selection and training of a machine learning model, based on the insights gained from the analysis of the data. Finally, the model is tested on new data, and the resulting inference is evaluated according to the most suitable among a wide range of metrics.

---

[23] *https://www.kasparov.com/ai-should-augment-human-intelligence-not-replace-it-harvard-business-review-march-18-2021/*

Let us explain further the aspects involved in the particular project of this thesis.

# 1. Data

Despite not properly constituting machine learning yet, this is perhaps the most crucial step in a program of this kind, as evidenced by its vital role in this project. As the popular saying "Garbage in, garbage out" summarizes, in fact, a program of this nature can only be as good as the data it is trained on. Microsoft researchers studied this exact aspect in 2001, concluding: *"these results suggest that we may want to reconsider the tradeoff between spending time and money on algorithm development versus spending it on corpus development."* (Banko & Brill, 2001). This is because raw data is the result of a chaotic and complex reality; hence, it inevitably tends to include noise, outliers, and inconsistencies.

Let us now dive deeper into the steps of this phase.

## 1.1. Data Collection

Data collection involves gathering the specific data needed for the machine learning project. Before this can be done, however, it is necessary to clearly identify the problem at hand, so that also the specificities of the required data are understood; only then, can one start looking for such data. As well as the type of data and its characteristics, this phase should also involve determining its size, that is the number of observations to be collected. In fact, whilst some machine learning models can adapt to small datasets rather well, it is not always the case, especially with more complex problems. Because such step has to, at least in part, take into account the model to be used later, it can be seen how machine learning projects tend to be partly iterative in their nature.

For the gathering of the data, multiple sources are often consulted and found; most of the time, it is either the case of premade databases or of online content, embedded in websites. In this latter case, web scraping techniques can be utilized, which involve automatically extracting data from websites using software tools or programs (as will be seen in the next chapter). Data might not always be there already, however, and more specific problems usually require tailored data to be measured and recorded in the first place, in fact. This was the case for the first of this project's tasks, in which case one is to collect information through surveys, interviews, or experiments.

Finally, the data needs to be stored and ordered in a way that suits the machine learning code that will follow, so that it can be extracted, read, and used by this latter. This is particularly crucial for data coming from different sources and/or in different formats.

During this data collection phase, data synthesis was also considered to address data imbalances and size limitations. This would involve generating new data based on existing data, often using statistical methods or machine learning algorithms, the most common methods being resampling, simulation, and augmentation. In the end, however, this strategy was discarded to ensure that the synthesized data was not mis-representing the original data and introducing biases or errors into the analysis. Instead, efforts were focused even more on data collection and preparation, described below.

## 1.2. Data Exploration

Once the necessary data has been obtained, it is time to make sense of it, so as to understand how to manipulate and prepare it for the machine learning model that will use it. This phase literally involves "exploring" the data at hand to gain insights about their attributes, formats, and general characteristics. Such process is usually carried out by making extensive use of statistics and visualization. The former, in fact, allows to investigate the range of the various variables, as well as useful measurements regarding their distribution, such as standard deviation and mean, metrics generally referred to as "summary statistics"; it can also help uncover correlations or other relationships among the different attributes of the dataset. Visualization, on the other hand, aids in presenting data in a format more easily understandable to us humans, which can make statistical patterns and trends including the ones described just above more evident or even immediate; it also tends to offer an initial, yet crucial idea of which model would best fit a particular dataset.

A vital function of data exploration lies in it also helping to identify potential issues with or biases in the data that could impact the accuracy and effectiveness of the model. Specifically, it offers insights into missing or erroneous data points, skewness or imbalances, outliers, and noise, setting the stage for the data preparation phase.

## 1.3. Data Preparation

This phase serves the purpose of cleaning, transforming, and manipulating the data so as to prepare it for its use in the machine learning model, and it stems directly from the discoveries made in the previous step.

**Data Cleaning**

To improve the quality of data and its suitability for analysis, it is of utmost importance that errors, inconsistencies, and inaccuracies are dealt with. In particular, outliers pose a significant threat to the algorithm correctly understanding the pattern followed overall by the dataset; these are data points presenting evidently distinct characteristics from the other records in the system. While many approaches exist to deal with outliers, the one adopted here has been to remove them by clipping the data at certain values, so as to better represent the overall distribution of the observations.

Another crucial part of data cleaning involved in this project has been the filtering and removal of irrelevant data from the set. As will be described in the next chapter, in fact, some observations were unrelated to the problem being studied and, therefore, would have distorted the predictions made by the model.

**Data Transformation**

No matter the data gathering process, it is likely the resulting data will require some form of feature engineering; that is, the process of selecting and transforming the raw input data into a set of features that can be used as input to a machine learning algorithm. This can include tasks such as selecting relevant variables, transforming variables, and creating new features based on domain knowledge. The most relevant of these in the problem at hand was feature selection: as will be discussed in detail later, in fact, not only were many attributes considered irrelevant or superfluous, but some were absent in several records, leading to potentially serious issues such as bias or general prediction errors. Overall, by reducing the numbers of variables, feature selection increases the efficiency and accuracy of the model, as well as its interpretability.

Afterwards, data scaling was applied to attempt better balancing the contribution of different variables in the dataset, although without particularly relevant results. This process involves transforming one or more of the features so as to have them all on a similar scale, which can help to ensure no single variable dominates the others, as well as to identify relationships and patterns in the data that may not have been apparent before scaling. The scaling applied in this case was a logarithmic one, which is useful when the distribution of the data is highly skewed and the majority of the values are concentrated in a narrow range, while a few extreme values are much larger or smaller than the rest. In such cases, in fact, logarithmic scaling can help to spread out the data and make the differences between values more comparable.

**Data Splitting**

Finally, data splitting was applied to ensure that the model be trained and evaluated on different subsets of the data, so that it can generalize well to new data points. As the standard dictates, a random selection process between train and test set was used, in order to avoid using the same data for multiple purposes and to ensure that the results are unbiased and reliable.

# 2. Model Selection and Training

By definition, a model is a simplified version of the true distribution of the observations, as it attempts to omit the superfluous details that are unlikely to generalize to new instances. In order to do so, however, it is forced to make assumptions. A linear model, for instance, assumes the data to be fundamentally linear and the discrepancy between the instances and the straight line to be mere random noise that can be safely ignored. In a notorious 1996 paper, David Wolpert demonstrated that if one makes absolutely no assumption about the data, then there is no reason to prefer one model over any other. Since there is no model that is a priori guaranteed to work better, then, the only way to determine the best one is to evaluate them all. Such principle took the evocative name of No Free Lunch (NFL)[24] theorem and it is now a core paradigm of the field.

This is precisely where the role of the data scientist comes into play: by looking at the data and drawing on their mental library of machine learning models, they are to select the most appropriate algorithms for a given problem and dataset. Next, by comparing the chosen models and their performances, they are to make the ultimate choice regarding which algorithm to use and proceed to its tuning and optimization, as well as testing and final evaluation. This back-and-forth interaction between the programmer and the program is yet another key example of the previously mentioned human-machine symbiosis which is more and more at the core of today's society.

Machine learning models come in many different types, based on different criteria: depending on whether they are trained with human supervision or not, they are typically distinguished into "supervised", "unsupervised", "semi-supervised", and "reinforcement learning"; if they can learn incrementally in real-time, they are referred to as "online-learning", as opposed to "batch-learning" or "offline-learning" models; finally, those working by simply comparing new data points to known data points are named "instance-based", whilst those detecting patterns in the training data and building a predictive model, similarly to a scientist, are "model-based". The two problems of this project both make use of supervised regression algorithms, learning offline and based on models rather than instances. Let us delve further into these specific aspects.

**Supervised Learning and Regression**

What distinguishes supervised learning from its contrary is the training data including the desired solutions, called *labels*. This means that, rather than having to learn to distinguish data points into categories by itself, the algorithm is already given guidance, through which it can focus on learning a model that fits the data or on classifying new data into the already given categories. Whether the algorithm has to perform such categorization or rather predict a target numeric value, it is said to perform *classification* or *regression*, respectively. This latter is the one of focus in this project.

Opposed to classification, as anticipated, regression attempts to deduce a precise number denoted as *target*, given a set of features called *predictors*. Examples of a task of this kind are predicting someone's income from variables such as their education or experience, as well as predicting one's grade based on factors along the lines of the hours spent studying and the hours of sleep before the test.

---

[24] *https://en.wikipedia.org/wiki/No_free_lunch_theorem*

**Batch Learning**

Despite the second program in this thesis making use of web scraping, it actually does the learning offline, without automatically updating as new data is added to the website, and the same is true for the first algorithm. The program could definitely be further developed to learn incrementally; however, due to the extremely low rate of data being added to the site, as well as to avoid potential connection issues from complicating the inner functioning of the code, the current version is completely valid as it stands. Furthermore, the program acts more as an "on-demand" software, which therefore updates its data each time it is used, in any case.

Downsides of batch learning can be the computing resources and time required, as the system needs to be trained from scratch each time new data is to be added. However, the greater simplicity resulting from such a program, as compared to one learning incrementally, can outweigh the drawbacks, as is the case for the program developed here.

**Model-based Learning**

Unlike instance-based learning, which memorizes the given examples by heart and compares the new data points to these, model-based learning attempts to capture the overall trend of the data through a function representing its behavior. The data scientist, in this regard, is to select one or more appropriate models for the given data as well as a performance measure to compare them, whilst it is then the models themselves which "fit" to the data, by trying several different parameters regulating factors like their shape and direction.

Following are the models involved in this study, individually described and explained. Most of them were employed in the second task of the project, yet also the first one briefly tapped into multiple of these tools.

## 2.1. Linear Regression

Linear regression (Figure 1) is, perhaps, the simplest model of all, yet also the most powerful. In fact, not only many relations can be explained by it, but its combination of simplicity and effectiveness have made it a dominant player in the field. Moreover, it is the basis of many more advanced models, which simply build on top of it to add complexity.

From a mathematical point of view, it essentially works by fitting a linear function to the given data. Such function consists of a weighted sum of the input features with an added constant called the *intercept* or the *bias*:

$$\hat{y} = \theta_0 + \theta_1 x_1 + \theta_2 x_2 + \cdots + \theta_n x_n,$$

which can be written in vectorized form as:

$$\hat{y} = h_\theta(x) = \theta \cdot x,$$

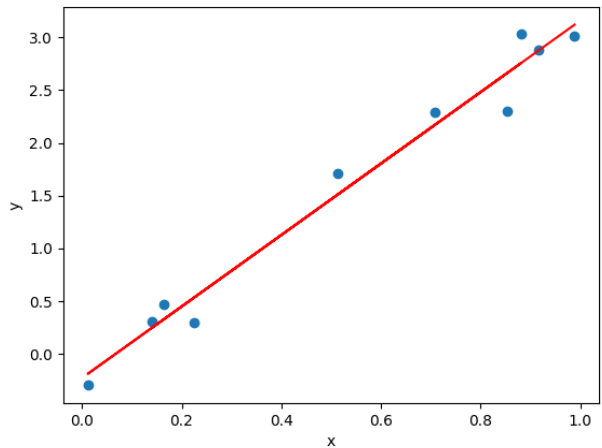

*Figure 1: Linear Regression*

where "$\hat{y}$" is the predicted value, n is the number of features, "$x_i$" is the $i^{th}$ feature value, and "$\theta_j$" is the $j^{th}$ model parameter. Accordingly, in the vectorized version, "**θ**" is the model's *parameter vector*, "**x**" is the instance's *feature vector*, and "**θ** · **x**" is the dot product of the vectors **θ** and **x**; "$h_\theta(x)$", instead, is the hypothesis function, which makes use of the model parameters **θ** and is applied on **x**. Two remarks to be made here are that: "$x_0$" is always equal to 1, as might have been noticed; whilst "$\theta_1, \theta_2, \ldots, \theta_n$" represent the feature weights, "$\theta_0$" is the bias term, hence the parameter vector **θ** contains both the bias term and the feature weights.

That is the linear regression model by itself. The way it is then trained is by finding the parameters that make it best fit the training set, according to a chosen performance measure, which usually represents either a measure of the error made by the model or a measure of its goodness of fit. In the former case, then, one is to minimize such measure, whilst in the latter they are to maximize it. Such optimization is usually performed either through an iterative approach, which gradually tweaks the model parameters to minimize the cost function over the training set, or through a direct "closed-form" equation, which directly computes such model parameters. This latter was the method utilized here, which makes use of the so-called *normal equation*:

$$\hat{\theta} = (X^T X)^{-1} X^T y,$$

where "$\hat{\theta}$" is the value of **θ** that minimizes the cost function, while "**y**" is the vector of target values. Through this formula, one is able to obtain the optimal parameters for the model straight away.

A remark regarding terminology has to be made here. When the relationship to be modeled is between the dependent variable "y" and a single independent variable "x", it is the case of a *simple* linear regression; in this scenario, the parameter vector **θ** only contains two entries, the bias term and one feature weight. When, on the other hand, two or more independent variables are present, *multiple* linear regression is at play, which is the one used in the first of the two models of this thesis.

## 2.2. Nonlinear Regression Models: Function-based

**Polynomial Regression**

The relationship between data points is often nonlinear, requiring an $n^{th}$ degree polynomial to be fit to them (Figure 2). By adding powers of each feature as new features, however, one can still make use of a linear model on this extended set of features to find the needed polynomial. The general form of polynomial regression would, therefore, be:

$$y = \beta_0 + \beta_1 x + \beta_2 x^2 + \cdots + \beta_2 x^m.$$

Because in this project a second-degree polynomial is used, however, the regression model reduces to the form:

$$y = \beta_0 + \beta_1 x + \beta_2 x^2.$$

Afterwards, the estimation of the coefficients is performed similarly to linear regression.

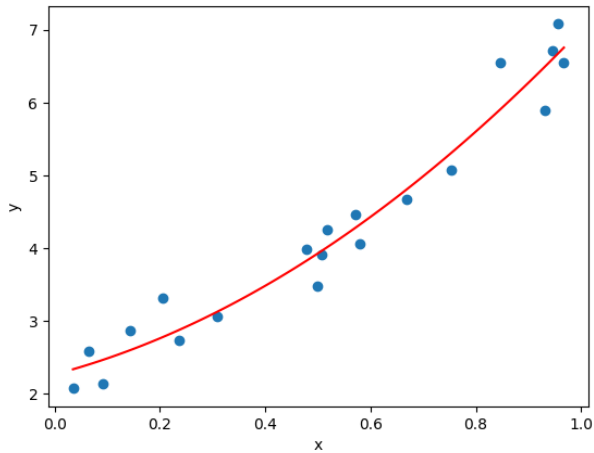
*Figure 2: Polynomial Regression*

**Logarithmic Regression**

As implied by the name, such model involves fitting a logarithmic curve to the data by passing the logarithm of the independent variable as an input to the linear regression model, rather than using the original independent variable itself:

$$y = a + b \cdot ln(x).$$

This is particularly useful to model situations where growth or decay initially accelerates rapidly but then slows down over time.

**Hyperbola-based Regression**

Finally, a hyperbola-based regression (Figure 3) fits the equation of a hyperbola to the data similarly to the procedure applied by the previous models:

$$y = \frac{\beta_1}{\beta_2 + X} + \varepsilon.$$

Whilst the two previous curves grew similarly, however, this one is sharper, as well as having an asymptote that the others lack. A necessary disclaimer is that this model ought not to be confused with hyperbolic regression, which is

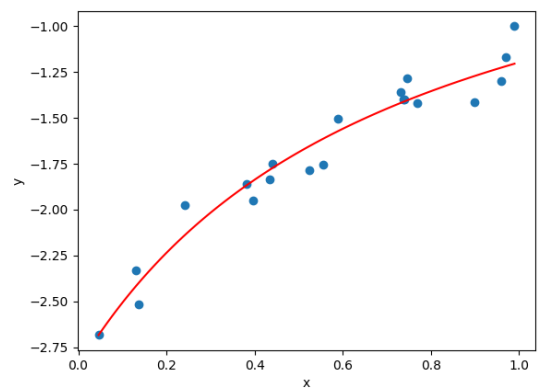
*Figure 3: Hyperbola-based Regression*

instead based on hyperbolic functions; these latter do derive from hyperbolas but are fundamentally different and will not be part of this discussion.

**Weighted Least Squares**

Weighted Least Squares (WLS) will also be implied as an attempt to account for heteroscedasticity, which is when the variance of the error terms in a regression model is not constant across all levels of the predictor variables. By giving different weights to observations based on one's specific needs, such issue can be countered, as it tends to "bend" the resulting function toward these more significant observations and less toward the others.

Overall, all these models aim to fit a nonlinear curve to the data, specifically one increasing at a diminishing rate and eventually leveling off. In fact, they will all be used in the second task of this thesis as alternative solutions to the problem. However, as will be seen later, their subtle differences will prove decisive in selecting the most appropriate model among them. Particularly, as anticipated above, what distinguishes them the most for our purposes is that polynomial and logarithmic curves do not actually have an asymptote they tend to, as opposed to the hyperbola one. This will make them increase indefinitely despite the data being capped at a certain value.

## 2.3. Nonlinear Regression Models: Tree-based

The following are some other algorithms to be used later which all stem from this first one, decision trees, and thus form a family of so called "tree-based" algorithms.

**Decision Trees**

Decision trees are algorithms partitioning the data space into smaller and smaller regions, based on the values of the input features. The resulting schema followed to partition such space resembles a hierarchical tree-like structure (Figure 4, left side), hence the name. The distance from the root (the first node) to the outmost leaf (any of the final nodes) is called "depth" and represents the number of decision nodes (splitting criteria) and leaf nodes. A tree too deep risks overfitting by capturing noise or idiosyncrasies in the training data that do not generalize well to new data, and vice versa for a tree too shallow; thus, such parameter will be adjusted during the analysis.

**Random Forests**

Random forests (Figure 4, right side) are so-called "ensemble" learning methods as they combine multiple models (decision trees, in this case) to improve the accuracy of the predictions. Each tree is trained on a different subset of the data and the final prediction is made by averaging the predictions of all the trees in the ensemble, the number of which is referred to as "number of estimators".

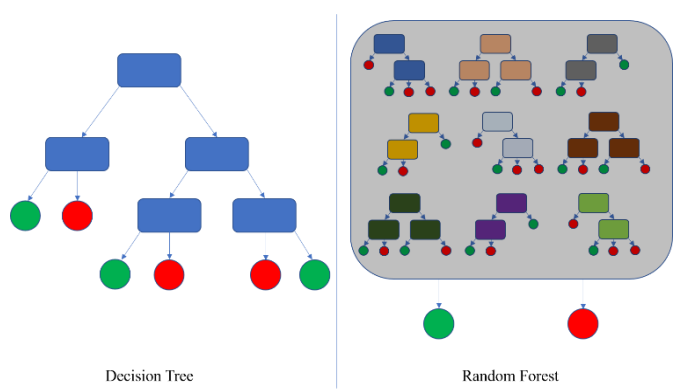

*Figure 4: Decision Trees and Random Forests*

**Boosting**

Boosting works similarly, yet each subsequent tree is trained to focus on the samples that were misclassified by the previous one, in order to improve the overall accuracy. The final prediction, then, is made by taking a weighted average of the predictions of all the individual trees. A remark to be made is that such technique is not limited to decision trees but can instead be applied to learners in general.

A particular version of this method is gradient boosting, which takes its name from the "gradient descent" [25] technique through which it optimizes the weights assigned to each individual learner. By complementing such feature with a variety of advanced techniques, the Extreme Gradient Boosting (XGBoost) algorithm has become one of the most popular and efficient algorithms in the field, including features as regularization, parallel processing, and tree pruning, the details of which will not be dealt with in this setting.

Now that the models to be used have been explained, it is time to look at their testing and validation methods, with particular care towards the performance measures to use.

# 3. Model Testing and Validation

The reason the data, as mentioned earlier, is split into two separate sets, the training one and the test one, is precisely for the purpose of evaluating how well the model generalizes to new cases. Whilst it is only trained on the first of these sets, in fact, the goal is actually for it to capture the underlying pattern that generalizes to all data of that kind. One needs to make sure the model has truly learned what it takes to perform well on this task no matter the precise data, rather than becoming proficient solely at this particular version of the task. If it only performs well on the training set, it is the case of "overfitting": whilst the training error is low, the so-called *generalization error* or *out-of-sample error* is high.

Let us, however, explore the various ways to measure this error in the first place; i.e., performance measures. Afterward, the resampling technique of cross-validation will be shown, which helps obtain a more reliable estimate of the model's generalization performance as compared to a single train-test split.

---

[25] *https://www.youtube.com/watch?v=IHZwWFHWa-w*

**Performance Measures**

Performance measures are part of the very essence of machine learning, as they are the parameters based on which models are evaluated and compared with one another; they are what actually testifies whether the desired learning has occurred or not, and in what proportion.

In regression problems as the ones here, the most common is the Root Mean Square Error (RMSE), which gives an idea of how much error the system typically makes in its predictions, with a higher weight for large errors. It is calculated as follows:

$$\text{RMSE}(\mathbf{X}, h) = \sqrt{\frac{1}{m} \sum_{i=1}^{m} (h(\mathbf{x}^{(i)}) - y^{(i)})^2},$$

with *m* representing the number of instances in the dataset, $\mathbf{x}^{(i)}$ being a vector of all the feature values (excluding the label) of the i$^{th}$ instance in the dataset, and $y^{(i)}$ being its label (the desired output value for that instance); then, $\mathbf{X}$ is a matrix containing all the feature values (excluding labels) of all instances in the dataset, and *h* is the system's prediction function, also called a *hypothesis*. In words, this formula squares the sum of the differences between the predicted values and the actual values, then takes the mean of such value, and finally takes its square root, so as to obtain an approximation of the average error made by the system.

Another popular measure is the Mean Absolute Error (MAE), which is less sensible to outliers, and is calculated as follows:

$$\text{MAE}(\mathbf{X}, h) = \frac{1}{m} \sum_{i=1}^{m} |h(\mathbf{x}^{(i)}) - y^{(i)}|.$$

A variation of this latter later used in the practical implementation is the Median Absolute Error (MDAE or MedAE), which simply uses the median value in place of the mean one in this same computation.

Finally, the coefficient of determination or R-squared ($R^2$) is a statistical measure representing the proportion of the dependent variable's variance that is explained by the independent variable. Because of this, it usually ranges from 0 to 1, where the latter represents a model perfectly fitting the data. At times, it can also yield negative values, for instance when linear regression is conducted without including an intercept or when a nonlinear function is used to fit the data. In fact, as will be seen in the second task of this project, it is not an appropriate measure for nonlinear models[26]. The R-squared is calculated as follows:

$$R^2 = 1 - \frac{RSS}{TSS},$$

where RSS is the sum of the squares of the residuals, while the TSS is the total sum of squares. In other words, the former measures the variation in the error between the observed data and modeled values, instead the latter measures how much variation there is in the observed data.

**Resampling Techniques: Cross-validation**

Cross-validation (Figure 5) is a resampling technique used to better evaluate the performance of a machine learning model, particularly when dealing with smaller datasets. The basic idea behind it is to simulate the process of training and testing the model multiple times on different subsets of the data, by partitioning it into several training and test sets and then averaging the results.

---

[26] Futher details about this at: *https://statisticsbyjim.com/regression/r-squared-invalid-nonlinear-regression/*

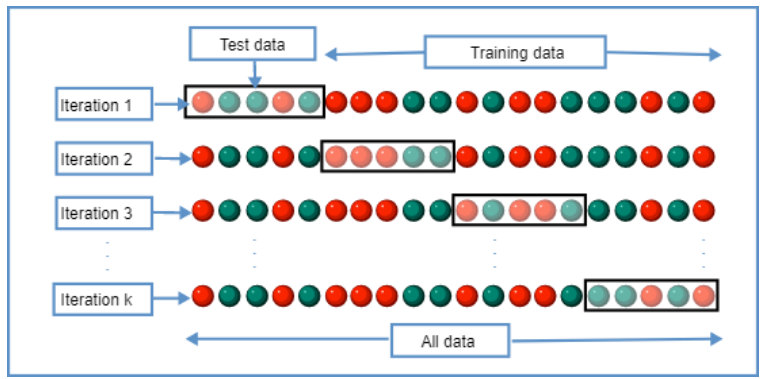

*Figure 5: Cross-validation*
*Creative Commons License: https://commons.wikimedia.org/wiki/File:K-fold_cross_validation_EN.svg*

Its most common form is k-fold cross-validation, which involves dividing the dataset into k equally-sized subsets or "folds". One of the folds is then held out as the validation set, while the remaining k-1 folds are used to train the model. This process is repeated k times, with each fold being used as the validation set exactly once. By averaging the performance of the model across the k validation sets, then, a more reliable estimate of the model's generalization performance is obtained, as compared to a single train-test split. This is because cross-validation makes use of all the data for both training and testing, while still maintaining a degree of independence between the training and validation sets.

A technique of this kind will be used in the first task of the project, due to the dataset's reduced size. In fact, cross-validation can be particularly useful in these cases, as it helps to make the most efficient use of the available data for model training and evaluation. In addition, it can also help to address some of the common challenges associated with small datasets, such as overfitting, model selection bias, and lack of statistical power. By using multiple folds and averaging the performance metrics across all the folds, cross-validation can provide a more robust estimate of the model's performance and help to identify potential sources of error or bias in the modeling process.

**Bias-variance tradeoff**

A fundamental theoretical result of statistics and machine learning is that a model's generalization error can always be expressed as the sum of three very different elements: bias, variance, and irreducible error.

The first is due to wrong assumptions in the nature of the data, resulting in a model too simple to capture the complex relationships between the data's inputs and outputs and therefore underfitting the training data. A typical instance of this is assuming the data to be linear when it is actually quadratic. The second, on the other hand, is due to the model's excessive sensitivity to small variations in the training data. A model with many degrees of freedom, for example, is likely to have high variance, and thus to overfit the training data. The last one, finally, is due to the noisiness of the data itself, the only way to reduce it being to clean up the data in the first place.

The ultimate goal in machine learning, thus, is to find a balance between bias and variance that minimizes the overall error of the model. The reason it is called a "tradeoff", in fact, is because increasing a model's complexity typically increases its variance and reduces its bias, while reducing a model's complexity increases its bias but reduces its variance (Figure 6). This will be seen on several occasions throughout the project explained in the next chapter.

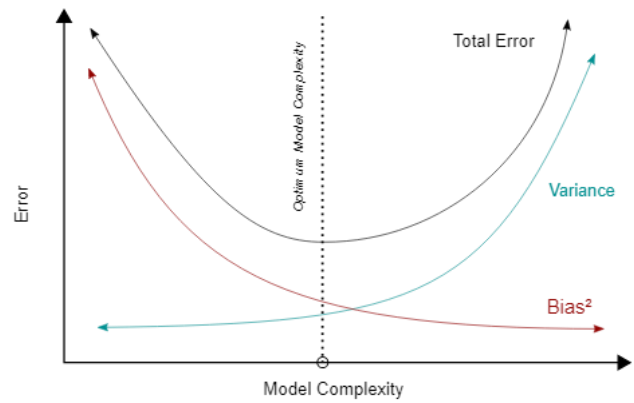

*Figure 6: Bias-variance Tradeoff*

*Creative Commons License:*
*https://commons.wikimedia.org/wiki/File:Bias_and_variance_contributing_to_total_error.svg*

# CHAPTER THREE

# PRACTICAL IMPLEMENTATION

As anticipated in chapter one, two distinct programs animate this research: one attempting to capture a universal relationship between an imperfect attempt at the 5-minute Numbers IAM discipline and its corresponding perfect attempt; the other trying to extract from an individual athlete's performance data the underlying curve representing their level at the Memory League Numbers discipline. Let us document the development of such software programs into the following sections of "Task I" and "Task II".

# *TASK I*

## 1. Data

As explained in the previous chapter, the quality of the data directly influences the results the machine learning program can achieve. In this first program, this aspect was in fact key because, while the relationship between the attributes was not particularly complex, the gathering and cleaning of the dataset required a much greater effort.

### 1.1. Data Collection

Let us start by identifying the data required. Upon finishing a 5-minute Numbers (or Speed Numbers) attempt on the IAM training software, the athlete is shown two different measures (Figure 7): the number of correctly recalled digits (in parenthesis) and the score (the larger number), which results from the penalization applied for the mistakes made. Other measures that could be considered, however, are: the total number of digits the athlete attempted to memorize, which differs from the number of correct digits by the amount of mistakes made (in this same picture, it would be 459); a short comment on the psychophysical state or on the feelings and reflections associated with that particular attempt, which is often recorded by the athlete in their training log. Whilst this latter was partly considered in evaluating the results analyzed, the number of total digits attempted was often missing, particularly because the software itself does not provide it; hence, while potentially useful for the analysis, it was discarded in favor of the number of correct digits, usually quite close to it, nonetheless. Together with it, of course, the score was considered, as the idea of this entire program was for it to bridge the gap between the score and the correctly memorized digits, by predicting an ideal quantity to aim at that usually sits in the middle of these two. Hence, to summarize, the two initial measures given by the software itself (that is, the score and the correct digits) were chosen for this task.

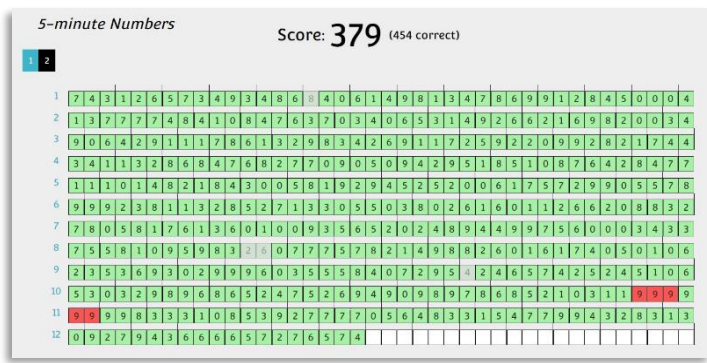

*Figure 7: Example Attempt at the 5-minute Numbers Discipline*

For the label, instead, the closest "perfect score" was needed for each attempt, meaning one representing what the ideal quantity to aim at would have been in the previous or successive imperfect attempt. Assuming progress takes time to occur, in fact, both the attempts right before and the ones right after a perfect attempt[27] can be thought to imply that same perfect score itself, meaning the athlete's underlying level has stayed relatively stable.

Figure 8 shows a typical training log of an athlete, to illustrate. As can be seen, the attempt on Friday was a perfect one, meaning the score (360) coincided with the quantity. Now, both the attempt right before and the one right after it seem to clearly indicate that a score of 360 could have been achieved by slowing down slightly, especially to the "trained eye" able to quickly interpret IAM scores. Thus, the pattern described just above of two imperfect attempts pointing to the same perfect one can be clearly observed.

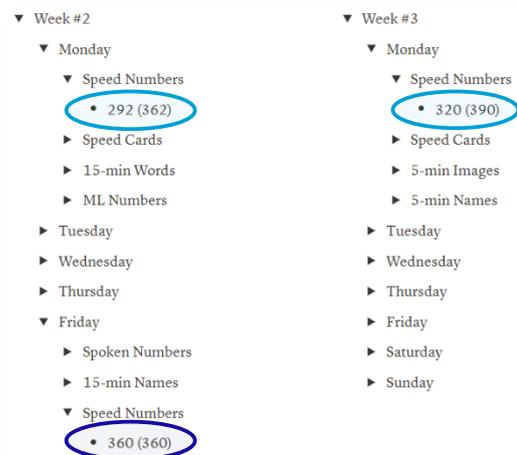

*Figure 8: Training Log with Perfect Score*

The gathering of the data involved a manual selection, from athletes' training logs, of such perfect scores, paired with their corresponding imperfect attempts. Such a procedure, of course, required domain expertise typical of a trained athlete used to making such estimations by themselves. At the same time, however, it avoided completely relying on the human component by having this latter only select pre-existing attempts and pairing them together, rather than also completely engineering such scores from scratch, as suggested later when referring to data augmentation.

Let us now further describe this gathering of the data. Machine learning needs data, and the more it has the more accurate are its predictions. Luckily, the algorithm applied was a linear regression rather than a more complex one, as will be seen; hence, the 133 records collected were enough for it to perform well. In fact, many attempts had to be discarded due to the lack of their corresponding perfect attempt, hence the resulting amount was severely reduced. Moreover, due to the manual selection described earlier, a larger amount of data would have required a much greater effort in selecting it. If truly needed, data augmentation techniques could have been applied, which in this case would have meant manually deciding for the label values of attempts initially discarded because of their lacking one; however, this is a risky procedure as it can lead to distorting the dataset and it also proved unnecessary, as already mentioned.

The data was collected from several athletes at varying levels of expertise and speed, providing a wide and representative range of observations. To ensure a more controlled supervision over the selection of such data and due to the peculiarity and partial complexity of the selection task itself, the data was requested in its raw form, to then be analyzed solely and directly by the author. Such selection, in fact, required both knowledge specific to the world of Memory Sports training as well as about the machine learning task to be performed with such data.

In the end, the gathered data appeared in the format shown in Figure 9, which was then converted into a CSV file (Figure 10) to be extracted later.

---

[27] One where the score corresponds to the correctly memorized digits because no mistakes were made.

| Score | CorrectData | SubsequentPerfectScore |
|---|---|---|
| 340 | 396 | 378 |
| 280 | 452 | 378 |
| 292 | 417 | 378 |
| 360 | 427 | 378 |
| 362 | 399 | 378 |
| 322 | 394 | 360 |

*Figure 6: Final Format of the Gathered Data*

```
Score,CorrectData,SubsequentPerfectScore
340,396,378
280,452,378
292,417,378
360,427,378
362,399,378
322,394,360
```

*Figure 10: CSV Version of the Gathered Data*

## 1.2. Data Exploration

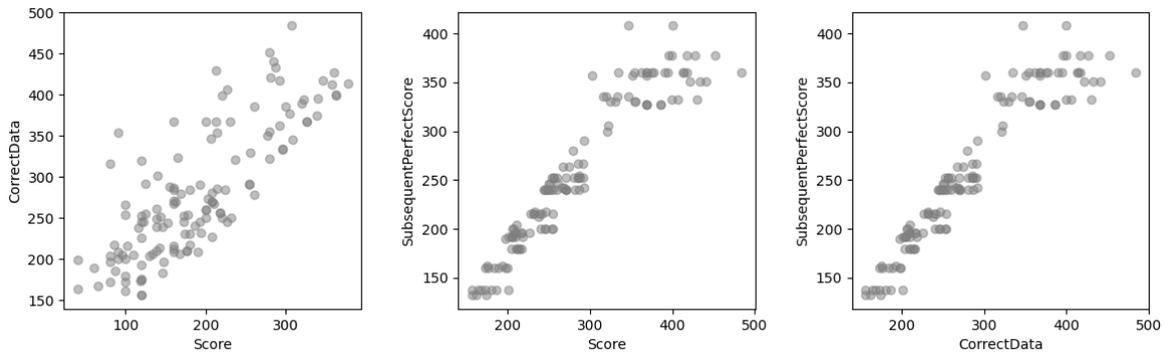

*Figure 7: Variables Plotted Against Each Other*

These are the variables plotted against each other (Figure 11). The score and the correct data have been named, respectively, "Score" and "CorrectData", while the label has been called "SubsequentPerfectScore", although it will also be referred to as "Score to Aim At", from here on. As shown, they all relate in a rather linear way, with some slight skewness towards the right, in the second and third case; this is likely due to higher memorization quantities attempted implying more errors and thus lower scores. A behavior of this kind, i.e. non-constant variance of a variable, is referred to as heteroskedasticity.

By looking at the same data in three dimensions, then, some outliers can be more easily spotted:

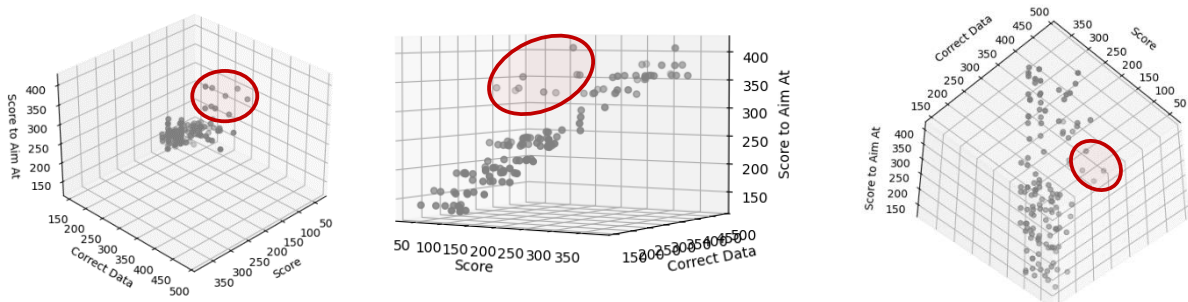

*Figure 8: 3D Plotting of the Variables Against Each Other, from Various Angles*

These will later be dealt with; nevertheless, the power of visualizing data in different forms proves decisive here, as the sole 2D representation of it was far from equally insightful.

Taking a look at their summary statistics (Code Excerpt 1), finally, we can see how the mean value, as well as all the percentiles, confirm the trend that will later be used to clearly identify outliers, that is the Perfect Score lying in between the other two variables, with some propensity towards the "CorrectData" one. This also makes intuitive sense, as the score usually drops substantially due to its penalizing nature; such fact is also confirmed by the "Score" column presenting the lowest minimum value of them all.

```
dataset.describe()
>>>
         Score    CorrectData  SubsequentPerfectScore
count  133.000000  133.000000              133.000000
mean   189.270677  280.383459              255.646617
std     80.069111   79.126742               75.755807
min     40.000000  156.000000              132.000000
25%    122.000000  214.000000              196.000000
50%    176.000000  260.000000              242.000000
75%    232.000000  347.000000              330.000000
max    377.000000  484.000000              408.000000
```

*Code Excerpt 1: Summary Statistics*

## 1.3. Data Preparation

This dataset did not require particular measures such as standardization or scaling, but it did require some cleaning and feature selection (Code Excerpt 2). As anticipated above, in fact, the total quantity of data memorized was not considered due to it missing in several records. This removed feature was also highly correlated to the one about correct data, as they only differed by a few units; hence, it could have also caused unnecessary complexity and distortion in the predictions. Moreover, perfect scores not in between the score and the correct data were removed, as they were rather rare and obsolete, and likely to just confuse the algorithm. Lastly, the data was split into a training and a test set:

```
mask = dataset['CorrectData'] < dataset['SubsequentPerfectScore']
dataset = dataset.drop(dataset[mask].index)
mask = dataset['SubsequentPerfectScore'] < dataset['Score']
dataset = dataset.drop(dataset[mask].index)
```

*Code Excerpt 2: Data Cleaning*

```
X = dataset.iloc[:, :-1].values
y = dataset.iloc[:, -1].values
X_train, X_test, y_train, y_test = train_test_split(X, y, test_size = 0.2, random_state = 0)
```

*Code Excerpt 3: Data Splitting*

Here is the data after cleaning and splitting:

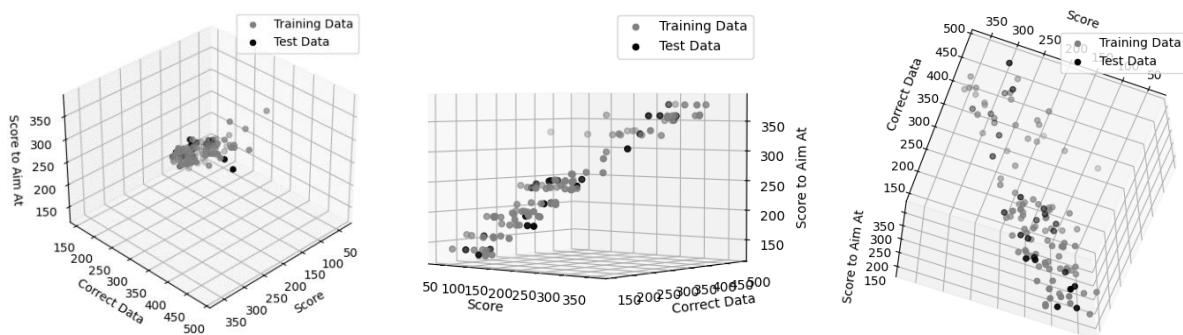

*Figure 93: Training vs Test Data*

## 2. Model Selection and Training

Due to the linear nature of the data and the overall simplicity of the task, the model selected was a multiple linear regression. The idea, in essence, was to fit a plane to the data (rather than a simple line), to be able to predict the perfect score based on the two input features of the score and the correct data of another attempt at the discipline. By plotting the cleaned features against each other just above (Figure 13), in fact, this clearly appears likely to be the appropriate function modelling their relationship. A simple line in 3D space might also appear to work, yet the plot on the left clearly shows the data to also be spread out laterally, rather than solely concentrated around a single axis; hence, the plane is the ideal candidate.

Other models were also tested: a regression decision tree and a random forest. Due to the simple relationship between the data and due to the limited dataset, however, these did not perform particularly better than the linear regression; what is more, as shown later, this latter performs rather well by itself. As explained in the previous chapter, in fact, unnecessarily complex models tend to overcomplicate things in these simpler scenarios, leading to worse results than more basic models. Moreover, interpretability of the model plays a crucial role in its selection; thus, due to the large difference in such property between the two types of models, the linear regression ultimately stands out.

```
regressor = LinearRegression()
regressor.fit(X_train, y_train)
```
*Code Excerpt 4: Linear Regression Training*

The model was trained using Scikit-learn's LinearRegression model on the training data (Code Excerpt 4). By plotting the resulting plane together with the *training* data, we can see how the two unravel together:

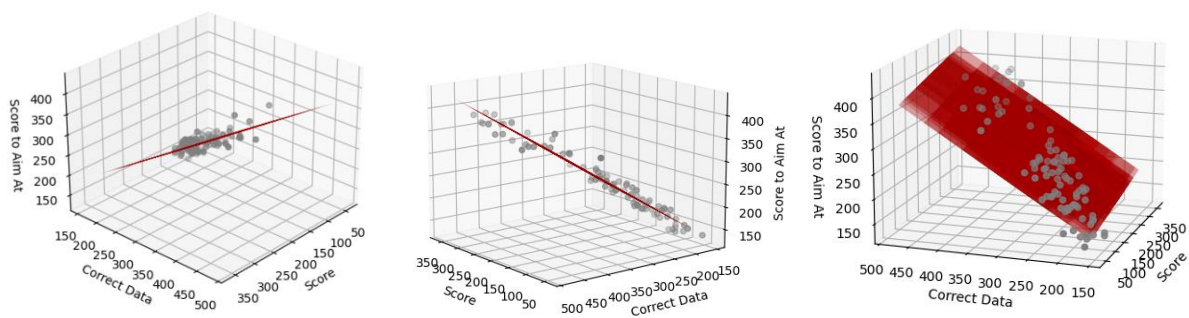

*Figure 14: Prediction Plane Against Training Data*

## 3. Model Testing and Validation

The resulting trained model was then used to predict the labels on the test set, which were rounded so as to avoid obtaining unnecessary errors in the predictions; in fact, since we know the data is discrete by nature, rather than continuous, we can

```
y_pred = np.round(regressor.predict(X_test))
```
*Code Excerpt 5: Label Prediction*

make the model only predict whole values in the first place (Code Excerpt 5).

This is how the plane appears together with the *test* data, this time:

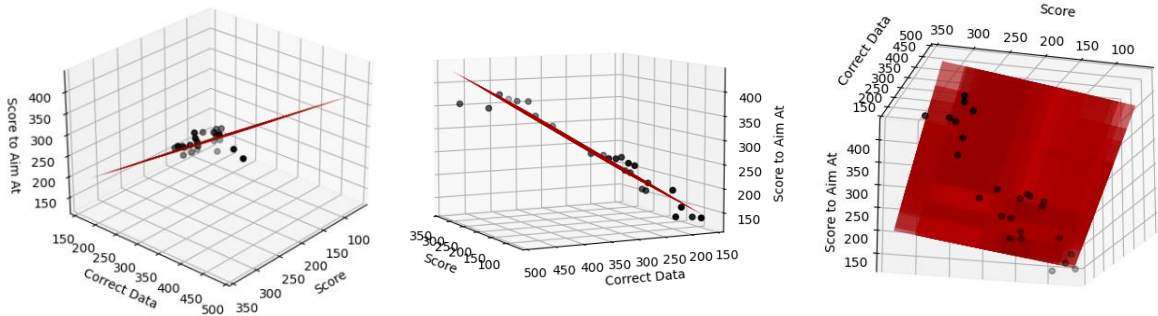

*Figure 15: Prediction Plane Against Test Data*

We can also obtain the precise parameters generating this specific plane by running Code Excerpt 6, obtaining Code Excerpt 7.

```
print("Intercept:", regressor.intercept_)
print("Coefficients:", regressor.coef_)
```
*Code Excerpt 6: Plane Parameters Computation*

```
Intercept: 11.843014003940112
Coefficients: [0.1897767  0.72575744]
```
*Code Excerpt 7: Plane Parameters Output*

Finally, the model was evaluated using several of the measures explained in the previous chapter, resulting in:

```
r2 = r2_score(y_test, y_pred)
mse = mean_squared_error(y_test, y_pred)
mae = mean_absolute_error(y_test, y_pred)
mdae = median_absolute_error(y_test, y_pred)
rmse = np.sqrt(mse)
```
*Code Excerpt 8: Performance Measures Computation*

```
R-Squared: 0.9283989367462151
Mean Squared Error: 429.52
Mean Absolute Error: 15.92
Median Absolute Error: 13.0
RMSE: 20.72486429388622
```
*Code Excerpt 9: Performance Measures Output*

The high R-squared indicates the model explains the variance of the data rather well, while the other measures suggest the deviation from the true values not to be too large; in fact, an error of 10-20 in this case is totally affordable and would be similarly made by the human expert, when asked to guess such measure.

This is the final plot, showing both the training and the test data, together with the prediction plane:

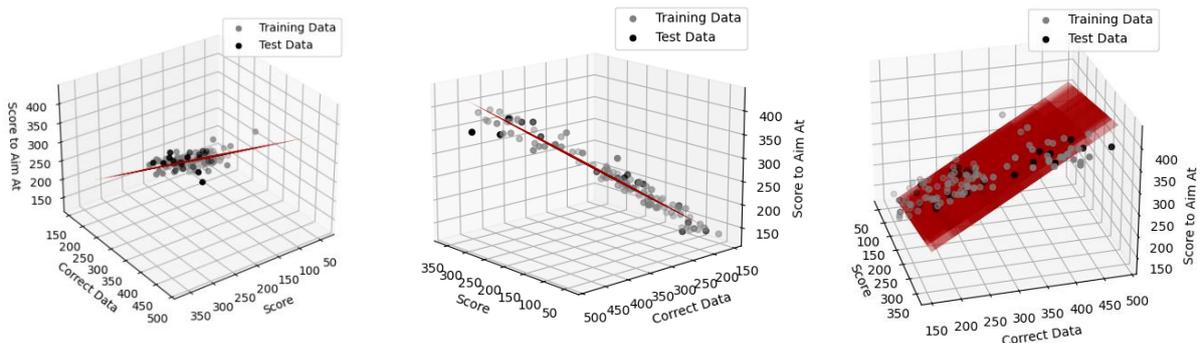

*Figure 16: Prediction Plane Against All Data*

the code to obtain which, and similarly all the previous plots, was:

```python
fig = plt.figure()
ax = fig.add_subplot(111, projection='3d')
ax.scatter(X_train[:,0], X_train[:,1], y_train, c='gray', marker='o')
ax.scatter(X_test[:,0], X_test[:,1], y_test, c='black', marker='o')
ax.set_xlabel('Score')
ax.set_ylabel('Correct Data')
ax.set_zlabel('Score to Aim At')
ax.legend(['Training Data', 'Test Data'])

# Plane
xx, yy = np.meshgrid(X_test[:,0], X_test[:,1])
zz = regressor.intercept_ + regressor.coef_[0] * xx + regressor.coef_[1] * yy
ax.plot_surface(xx, yy, zz, color='red', alpha=0.1)

plt.show()
```

*Code Excerpt 10: Graph of the Prediction Plane against both Training and Test Data*

**Cross-validation**

Because of the reduced size of the dataset, cross-validation appeared as a useful additional method to apply to check the validity of the results obtained. As anticipated in the previous chapter, in fact, this technique helps avoid common issues associated with small datasets by providing a more robust estimate of the model's performance, as it makes the most efficient use of the available data. In particular, it prevents one from considering the results associated with a particular split of the dataset as representative of any of its possible splits.

After setting a "k" for the number of folds in the cross validation, the different scoring measures were calculated:

```python
r2_scores = cross_val_score(regressor, X, y, cv=k, scoring='r2')
mse_scores = cross_val_score(regressor, X, y, cv=k, scoring='neg_mean_squared_error')
mse_scores = abs(mse_scores)
rmse_scores = np.sqrt(mse_scores)
mae_scores = cross_val_score(regressor, X, y, cv=k, scoring='neg_mean_absolute_error')
mae_scores = abs(mae_scores)
medae_scores = cross_val_score(regressor, X, y, cv=k, scoring='neg_median_absolute_error')
medae_scores = abs(medae_scores)
```

*Code Excerpt 11: Cross-validation Comparison of Performance Measures*

resulting in these outcomes:

| k | R-squared | MSE | RMSE | MAE | MedAE |
|---|---|---|---|---|---|
| 2 | 0.896433 | 473.845731 | 21.714649 | 16.472297 | 14.74841 |
| 3 | 0.890865 | 428.320277 | 20.629927 | 16.013775 | 14.388888 |
| 4 | 0.852901 | 368.941433 | 19.076528 | 14.730803 | 12.772012 |
| 5 | 0.838191 | 342.556472 | 18.188595 | 14.305168 | 13.284262 |
| 6 | 0.725104 | 346.272385 | 17.862046 | 14.578946 | 13.8717 |
| 7 | 0.682918 | 313.379851 | 17.047861 | 13.748409 | 11.920436 |
| 8 | 0.658537 | 323.618369 | 17.53959 | 13.934397 | 12.84475 |
| 9 | 0.720143 | 320.711399 | 17.140865 | 13.975427 | 12.735671 |

*Table 1: Heatmap of Cross-validation Scores Comparison*

Here is a graphical representation of the same results:

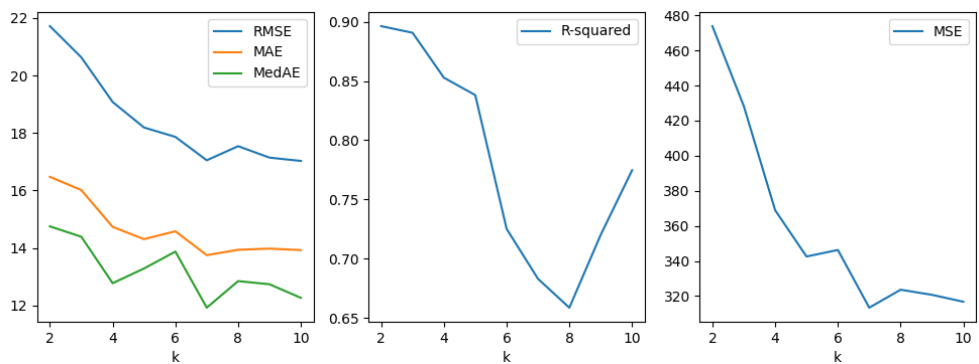

Figure 17: Cross Validation Results

The plots had to be separated because these measures have different scales. However, Figure 18 shows a single plot making use of standardization, to better compare the behavior of the measures as a whole.

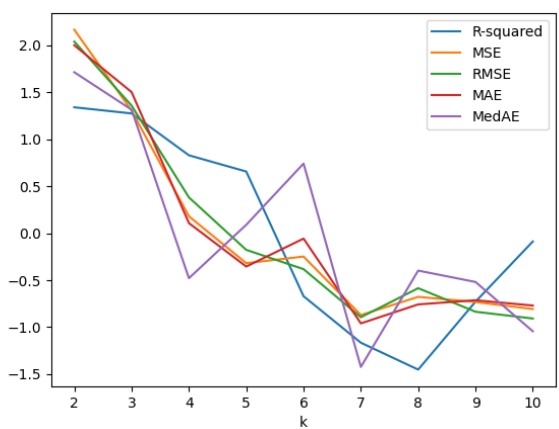

Figure 18: Standardized Comparison of the Cross-validation Results

The results show the initial model was already a fairly accurate representation of the situation, yet a clear pattern emerges and requires further explanation: as the number of K folds increases, all performance measures go down; this is because, despite each fold becoming smaller, the overall amount of data used for both training and testing increases. Therefore, the cross-validation estimate becomes less dependent on a specific split and is more representative of the overall performance of the model. Out of these measures, however, also the R-squared decreases, which is usually not a good sign: this stems from it being a measure of how well the model fits the data; since an increase in K results in each fold being smaller, the model has less data to learn from during each training phase. When this is the case, the model may be less able to capture the true underlying relationships between the independent and dependent variables, resulting in a weaker fit to the data. For values of k larger than 8, however, this same measure appears to grow again: a possible explanation is that the model is now starting to overfit the data, as it is prompted with excessively small datasets. In the end, the cross-validation appears to confirm the initial results for the performance measures as fairly accurate.

# TASK II

## 1. Data

Similarly to the first one, also this second program needed a certain amount of work to be put into the data; on top of that, however, this prediction problem was more complex.

### 1.1. Data Collection

As a reminder, the aim here is to extract from an individual athlete's performance data the underlying curve representing their level at the Memory League Numbers discipline. After an attempt at the discipline, the athlete is shown the number of correctly memorized digits (out of the maximum of 80) and the time it took them; this format, in fact, also rewards one for finishing earlier, as long as they obtained a score at least as high as their opponent. Being a head-to-head competition platform, Memory League also gives the chance to play against other athletes, in which case the results are shown next to the opponent's. In such one-vs-one matches, players are also shown the percentage of correct data to the total data, but that is a derived attribute that will not be needed for our purposes. Furthermore, in the case of a rated match, rather than a friendly one, athletes are also given respective changes in their ranking points resulting from the match; however, this factor does not truly impact our analysis, and hence will not be considered. Figure 19 pictures an example attempt at the discipline in solo mode, while Figure 20 shows a concluded match between two players.

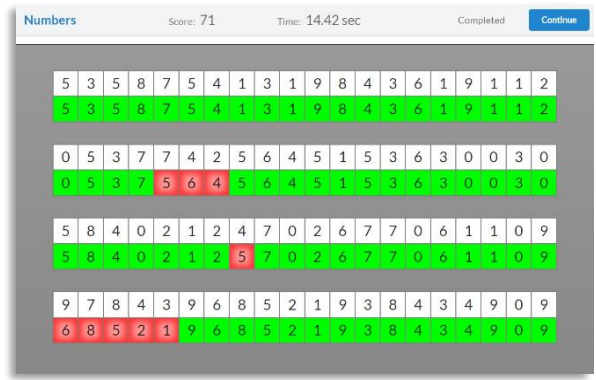

*Figure 109: Solo-mode Attempt at Memory League Numbers*

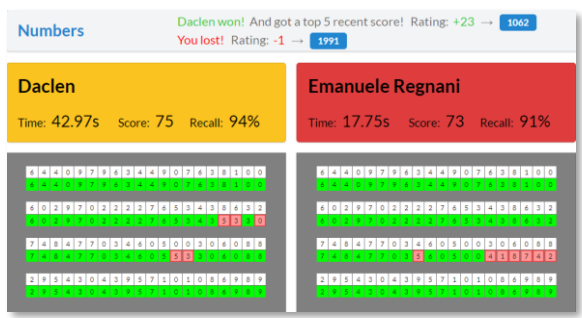

*Figure 20: Head-to-head Match in Memory League Numbers*

A first decision to make is which attempts to consider for the data gathering among the ones done in single-player mode, head-to-head, or both. The choice here was to solely consider the rated head-to-head matches, for two reasons. First of all, for a matter of data availability and overall flexibility of the program: attempts done in solo mode, in fact, would have required a manual process of reaching out to individual athletes to be collected, similarly to the one involved in the first program of this thesis. However, being this program centered on single athletes rather than on a general principle of memory training, even hundreds of data points from a particular practitioner would have only sufficed to train the model related to their own personal performance, rather than allowing it to adapt to different athletes. Moreover, being this discipline shorter than the 5-minute one covered earlier, not all attempts are even recorded at all and training tends to be slightly more experimental and random, thus perhaps less representative of the performance level shown in head-to-head matches. As anticipated, then, the second reason to not consider solo trainings was to capture attempts more reflective of the performance shown by athletes in actual rated matches against an opponent. Unlike when on their own, in fact, players tend to express their best potential when challenging someone else. Consequently, this reduces the noise in the data and also avoids potential incomplete or even illicit attempts that might appear in single-player training.

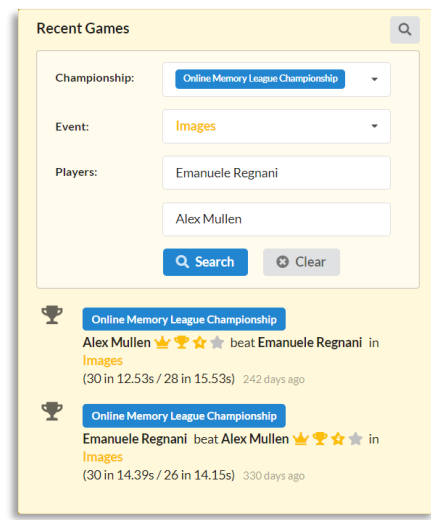

*Figure 21: Memory League's Search Filter for Head-to-head Matches*

The solution adopted, instead, was to use web scraping to extract the results of matches directly from the website. Rated matches, in fact, directly appear on the platform and can even conveniently be searched and filtered (Figure 21). This specific solution was also adopted so as to obtain more up-to-date information, as the data is taken each time from the current version of the website. Nevertheless, since players only play a few matches per month, this would not have impacted on the results too significantly. A final reason to use this method, however, was for its incorporated filtering and searching functionalities shown earlier. Upon receiving an athlete's name and a discipline as inputs, in fact, the program is able to find the corresponding matches on the platform and collect them in a text file (note: the search fields are optional, so even just matches involving one particular player in no particular tournament can be found, which is the case here). Here follows the code for the web scraping:

```
### Collect the input and open the website
athlete = input("Enter an athlete's name: ").title()
parsed_athlete_name = [*athlete]
discipline = input("Enter a discipline: ").title()
disciplines = {
    'Numbers': 'div.transition:nth-child(4) > div:nth-child(6)',
    'Cards': 'div.transition:nth-child(4) > div:nth-child(2)',
    'Words': 'div.transition:nth-child(4) > div:nth-child(7)',
    'Images': 'div.transition:nth-child(4) > div:nth-child(3)',
    'International Names': 'div.transition:nth-child(4) > div:nth-child(4)',
    'Names': 'div.transition:nth-child(4) > div:nth-child(5)',
    }
discipline_id = disciplines[discipline]

browser = webdriver.Firefox(executable_path=r'C:\Users\emanu\Programs\geckodriver-v0.31.0-win64\geckodriver.exe')
browser.get('https://memoryleague.com/#!/home')

### Search for athlete's games in the specified discipline
```

```python
button = browser.find_element(By.CSS_SELECTOR, "#gameSearch")
button.click()
sleep(0.5)
button = browser.find_element(By.CSS_SELECTOR, ".eventName")
button.click()
sleep(0.5)
button = browser.find_element(By.CSS_SELECTOR, discipline_id)
button.click()
sleep(0.5)
pyautogui.click(938,286)
sleep(0.5)
parsed_athlete_name.append('Enter')
pyautogui.typewrite(parsed_athlete_name)
parsed_athlete_name.pop()
sleep(2)

### Loop going through the search results
with open('Project/Dataset/' + athlete + ' - ' + discipline + '.txt', 'w') as f:

    ranges, x = [(1, 7)], 7     # (search results are displayed 7 at a time)
    while x < 400:
        ranges.append((x, x+10))
        x += 10

    for r in ranges:
        for i in range(r[0], r[1]):
            user_0 = 'a.publicMatchLink:nth-child(' + str(i) + ') > div:nth-child(2) > div:nth-child(1) > span:nth-child(2)'
            user_1 = 'a.publicMatchLink:nth-child(' + str(i) + ') > div:nth-child(2) > div:nth-child(1) > span:nth-child(6)'
            text = 'a.publicMatchLink:nth-child(' + str(i) + ') > div:nth-child(2) > div:nth-child(1) > span:nth-child(10)'

            if browser.find_element(By.CSS_SELECTOR, user_0).get_attribute("innerHTML") == athlete:
                result = browser.find_element(By.CSS_SELECTOR, text).get_attribute("innerHTML")[20:32]
            else:
                result = browser.find_element(By.CSS_SELECTOR, text).get_attribute("innerHTML")[52:-20]

            if result[-1] == 's':
                result = result[:-1]

            f.write(result[:3].strip() + ',' + result[-6:].strip() + '\n')
            print(result[:3].strip() + ',' + result[-6:].strip())

        # click "View more"
        button = browser.find_element(By.CSS_SELECTOR, 'div.message:nth-child(2) > p:nth-child(7) > a:nth-child(1)')
```

```
        button.click()

        # Scroll down to the bottom
        sleep(1)
        htmlelement= browser.find_element(By.TAG_NAME, 'html')
        htmlelement.send_keys(Keys.END)
```

*Code Excerpt 12: Web Scraping*

Finally, the data is automatically saved in the text file format of Figure 22, with the values for quantity memorized and time taken separated by commas.

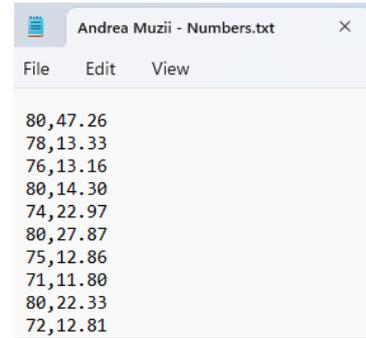

*Figure 22: Web Scraping Results*

Before moving onto the exploration of the data, it should be pointed out that this program, unlike the previous one, treats an athlete's data as representing a similar level of skill. For athletes having improved remarkably in the last months therefore (which is sometimes the case), older data would not properly reflect their current level, and hence the entire program would not prove particularly valuable. However, not only is this a minority in the world of Memory Sports, but these athletes would also likely not be that prone to accurate measurements anyway, due to their instability in the new ranges achieved; even more importantly, a program of this kind would probably be most useful to experienced athletes battling others on a matter of seconds of difference, as it is them who truly need to care about the details and precision provided by a program of this kind. Hence, this first version of the software will serve primarily for the purposes of these latter, yet a future version for the rest of the audience is not to be excluded and will partly be discussed in the concluding chapter.

## 1.2. Data Exploration

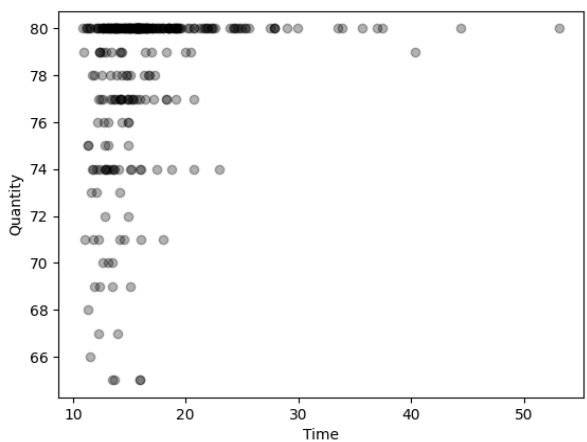

*Figure 23: Raw Data*

Taking IAM World Champion Andrea Muzii's data for the Numbers discipline (Figure 23), as an example, a clear nonlinear trend appears (and the same goes for other athletes). As anticipated in the first and second chapters, it appears to be a curve increasing at a diminishing rate and eventually leveling off at the maximum quantity possible in Memory League, i.e. 80 digits. The curve also plateaus vertically on the left, at the minimum time ever attempted by the athlete for that specific quantity of data. This asymptotic behavior of the ideal curve will prove crucial in the selection of the best model for this type of dataset.

Something worth noting is the discreteness of the values: only whole numbers appear on the Y axis, a detail that will prove valuable when predicting these values ourselves. Another particular characteristic to notice, then, regards the varying density of data points: despite the curved shape of the overall trend, a striking majority of attempts lie at the top of the graph, independently of the X axis. This is a clear sign of the athlete attempting to achieve the maximum quantity of 80 digits no matter the time taken. In fact, the general pattern shown in the plot can simply be described as the athlete's accuracy declining as they take less and less time for the memorization. This diversity in density will also be crucial in the model training phase, as it will influence the curvature of the function much more than the human eye would suggest. Furthermore on this density topic is the greater presence of 77s and 74s: while this will not be particularly significant in our analysis, it has a clear explanation to the trained eye. This and many other advanced athletes, in fact, memorize chunks of three digits at a time, making it more likely to have gaps in memory of this same particular size as compared to any other; therefore, gaps of one (77 instead of 80 digits) or two (74 instead of 80) chunks appear more frequently in the data than those of just any number of single mistaken or forgotten digits.

```
           Quantity        Time
count    304.000000  304.000000
mean      78.088816   16.526151
std        3.336270    5.245910
min       65.000000   10.820000
25%       77.000000   13.497500
50%       80.000000   15.240000
75%       80.000000   17.512500
max       80.000000   53.190000
```

*Code Excerpt 13: Summary Statistics*

In terms of summary statistics, of course, the data presents an accentuated skewness towards quantities of 80 and times below 20 seconds, as shown by the quantiles (Code Excerpt 13). Naturally, this is due to the athlete attempting to achieve a perfect score of 80 each time, whilst on the other hand trying to keep the time taken as low as possible.

Similar considerations can emerge by viewing a histogram of each of the two variables (Figure 24).

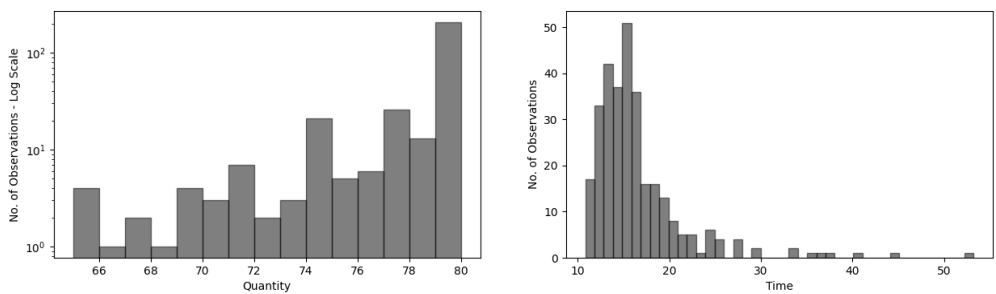

*Figure 24: Histogram of the Distribution of Andrea Muzii's Data*

## 1.3. Data Preparation

As mentioned above, several variables were excluded from the analysis: apart from the previously mentioned ranking points and percentage of correct data, also the specific competition in which a match happened and the opponent faced were discarded; this was decided because such features would not only unlikely have had a relevant impact on the results, but they would have also introduced unnecessary complexity in the model. A player might employ different strategies based on their opponent or on the type of competition, but their underlying skill level would not change and that is precisely what is being analyzed here. In other words, we are not trying to determine whether they usually "go slow" or "fast" or similar parameters, but rather what score they tend to get in certain time ranges, and vice versa, independently of whether they consciously chose to aim for a certain speed or not. The greater variability resulting from the different speeds reached in various occasions, actually proves helpful, then, as it adds variation to the dataset by covering more of the total spectrum.

In preparing the dataset for the model training, an attempt was made to scale the time variable logarithmically; because such power-law behavior was only modest, however, such attempt did not prove particularly useful and was thus abandoned. Moreover, as will be explained later, the logarithmic curve itself was not what was needed in this particular scenario.

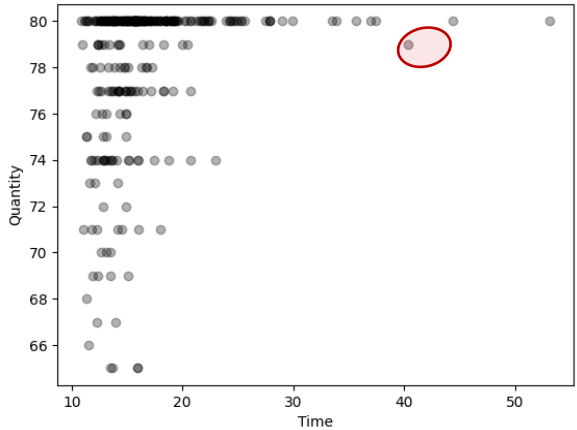

Figure 25: Outlier in Andrea Muzii's Data

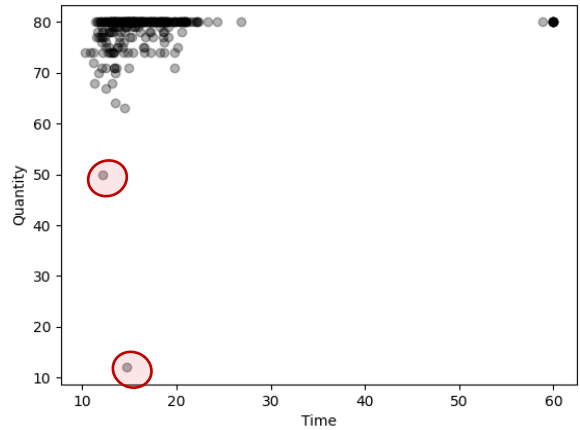

Figure 26: Outliers in Alex Mullen's Data

```
mask1 = data.iloc[:, 0] != 80
mask2 = data.iloc[:, 1] > np.percentile(data.iloc[:, 1], 95)
data = data.drop(data[np.logical_and(mask1, mask2)].index)

mask = data.iloc[:, 0] < np.percentile(data.iloc[:, 0], 100-99)
data = data.drop(data[mask].index)
```

Code Excerpt 14: Data Cleaning

Moving onto data cleaning, (Code Excerpt 14) outliers first needed to be dealt with: Figure 25 shows a clear example, as the athlete made a mistake despite the time taken to memorize being rather high for their standards. To avoid such extreme cases influencing the predictions, values exceeding the 95th percentile and being smaller than 80 were excluded from the dataset. A similar measure also had to be taken for outliers in the Y direction, as demonstrated by other athletes' datasets; for instance, Figure 26 shows Memory League World Champion Alex Mullen's data.

A particular measure that was attempted was to temporarily take out part or even all the 80s (perfect scores) from the dataset, training the model solely on the data points from 79 downward. The idea was to avoid the exaggerated influence of the extremely dense cluster formed by the data at the top to better focus on capturing the pattern happening below it; however, this approach was then abandoned upon realizing that the 80s were actually an integral part of the overall trend, constituting most of its horizontal stretch and giving it its characteristic exponential shape, in fact.

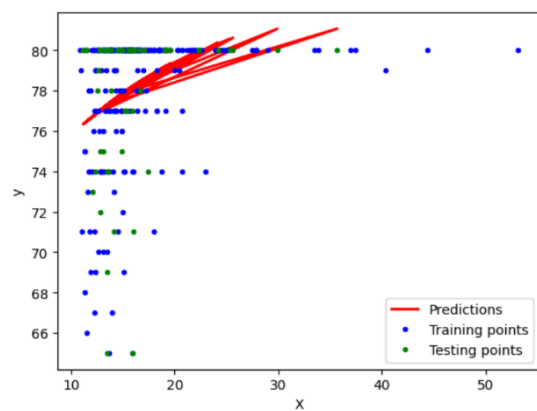

Figure 27: "Chicken Crest" Behavior

```python
def myownsplit(X, y):
    X_train, y_train, X_test, y_test = [], [], [], []
    
    for i in range(len(X)):
        if i%5 == 0:    # 1/5 of the data for testing
            X_test.append(X[i])
            y_test.append(y[i])
        else:
            X_train.append(X[i])
            y_train.append(y[i])
    
    return X_train, y_train, X_test, y_test
```

*Code Excerpt 15: Custom-made Splitting Function*

Finally, the data was split into a training and a test set, similarly to Task I of this project. However, for the function-based models, a special splitting function had to be defined to maintain the order of observations relative to each other (Code Excerpt 15). Without it, in fact, the sequential nature of the models would result in them exhibiting this "chicken crest" behavior (Figure 27) instead of plotting a clean curve, because they would sequentially attempt to connect the unordered points with a line.

## 2. Model Selection and Training

Model selection was much more challenging for this task. Initially, in fact, the natural choice seemed to be a polynomial curve resembling the exponential distribution of the data, yet things did not play out that simply.

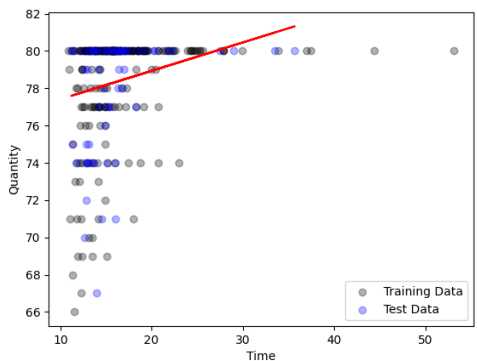

*Figure 118: Simple Linear Regression*

```
regressor = LinearRegression()
regressor.fit(X_train, y_train)
y_pred = regressor.predict(X_test)
```

*Code Excerpt 16: Linear Regression Training and Prediction*

For starters, a simple linear regression was taken as the baseline model (Code Excerpt 16, Figure 28), a benchmark against which to compare the results of all the other models. As the data clearly does not exhibit such linear behavior, in fact, any model performing

```
poly = PolynomialFeatures(degree=(2,2))
X_train_poly = poly.fit_transform(X_train)
X_test_poly = poly.transform(X_test)
poly_reg = LinearRegression()
poly_reg.fit(X_train_poly, y_train)
y_pred = poly_reg.predict(X_test_poly)
```

*Code Excerpt 17: Polynomial Regression Training and Prediction*

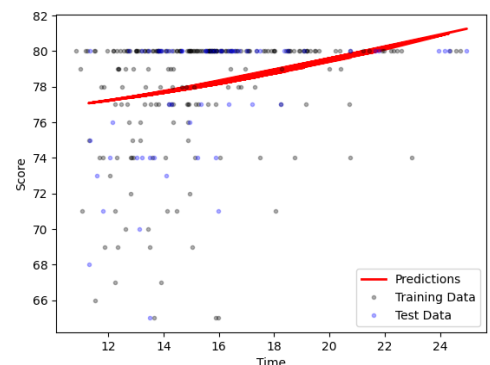

*Figure 29: Polynomial Regression*

worse than this is undoubtedly inappropriate for the task.

Upon testing a polynomial model (Code Excerpt 17), however, an even worse curve was found, curling in the wrong direction, upward (Figure 29).

The immediate response, then, was to impulsively go for more complex models. Firstly, a decision tree was used (Code Excerpt 18).

```
model = DecisionTreeRegressor(max_depth=10)
model.fit(X_train, y_train)
y_pred = model.predict(X_test)
```

*Code Excerpt 18: Decision Tree Training and Prediction*

Encouraged by the results, further effort was put into this direction by training a model making use of multiple trees: a random forest (Code Excerpt 19). The model, however, did not perform well, as will be shown in the final section of this chapter.

```
from sklearn.ensemble import RandomForestRegressor
regressor = RandomForestRegressor(n_estimators = 1000, random_state = 0)
regressor.fit(X_train, y_train)
y_pred = regressor.predict(X_test)
```

*Code Excerpt 19: Random Forest Training and Prediction*

As a final attempt on this track, nonetheless, Extreme Gradient Boosting (XGBoost) was tested:

```
xgb_model = xgb.XGBRegressor(objective="reg:squarederror", random_state=42)
xgb_model.fit(X_train, y_train, eval_set=[(X_test, y_test)], early_stopping_rounds=10, verbose=2)
y_pred = xgb_model.predict(X_test)
```

*Code Excerpt 20: XGBoost Training and Prediction*

In general, by looking at the way the splits are made in these tree-based models (Figure 30), it appears as if the algorithm is attempting to learn the relationships by heart, extracting principles such as *"if the time taken is between a and b, then the quantity must be c"*, overfitting to the training data. Moreover, the limited interpretability and with it the difficulty in comparing different athletes together posed an issue.

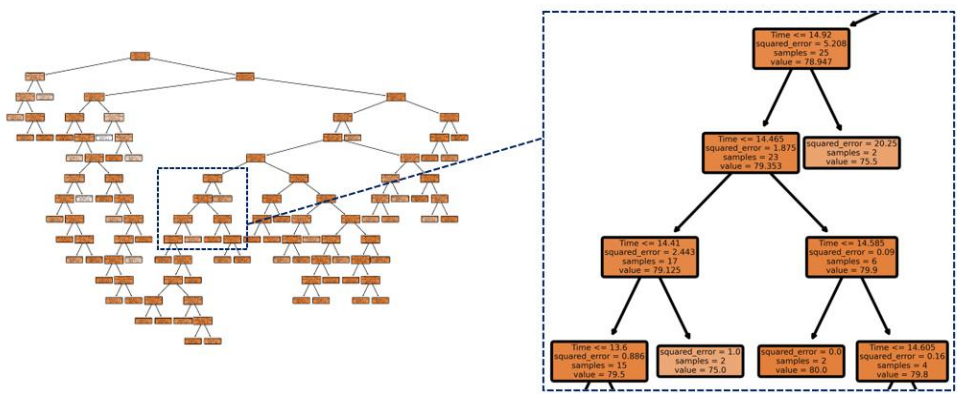

*Figure 30: Tree-based Model Example*

Suspicion arose about not having fully explored the initial approach of a polynomial curve, as no clear reason for it not working properly had been uncovered. Realizing the data exhibited a behavior extremely akin to a logarithmic function, then, this latter was tested in hope of obtaining better results (Code Excerpt 21).

```
x_log = np.log(X_train)
model = LinearRegression()
model.fit(x_log.reshape(-1, 1), y_train)
x_test_log = np.log(X_test)
y_pred = model.predict(x_test_log.reshape(-1, 1))
```

*Code Excerpt 21: Logarithmic Regression Training and Prediction*

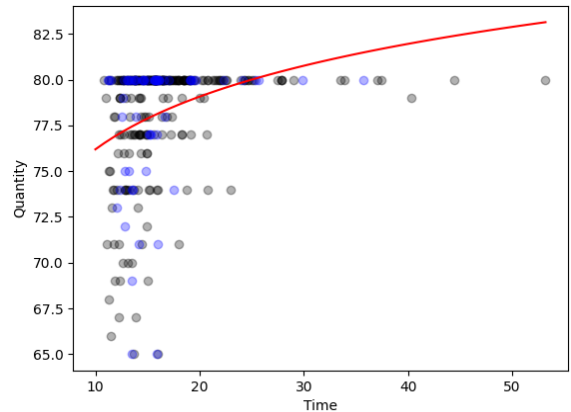

*Figure 31: Logarithmic Regression*

Although not competitive at all with the tree-based models and still far from optimal, it seemed to be the right path to go down. The main issue appeared to be the curvature of the function not being sharp enough and instead "opening up" excessively (Figure 31).

```
weight = 1/y**8
poly = PolynomialFeatures()
reg = LinearRegression()
x_poly = poly.fit_transform(x)
reg.fit(x_poly, y, weight)
y_pred = reg.predict(x_poly)
```

*Code Excerpt 22: Weighted Least Squares Training and Prediction*

Thus, the particular approach of weighted least squares regression was adopted (Code Excerpt 22) as it seemed the 80s at the top were having an exaggerated effect on the model by "dragging" the curve upward. By having data points with lower Y values weigh more in the regression, in fact, the overall curve could be brought down to possibly better reflect the underlying trend.

Similarly to the attempt made in the data cleaning part to completely remove these data points at the top, however, this approach proved faulty as these high values were actually crucial to the prediction process. The curve, in fact, started at a sufficiently low value, but then overshot and even came back down, completely missing the trend in the end (Figure 32). An improvement was obtained by cutting off the outer part of the dataset only containing 80s, focusing on the area with the most variation in the observations (Figure 33); nevertheless, the curve still blundered due to its fundamental disregard for the 80s at the top, which instead clearly appear to play a role in shaping the ideal curve.

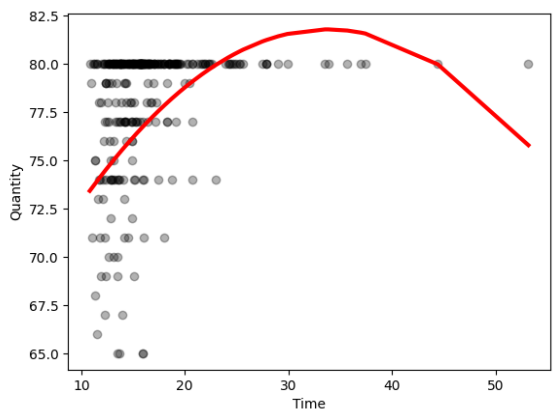

*Figure 32: Polynomial Weighted Least Squares Regression*

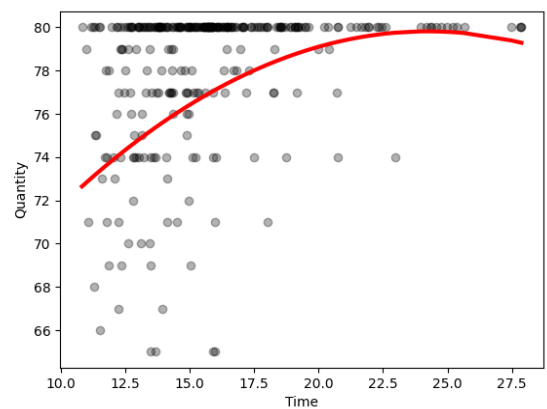

*Figure 33: Cut-off Polynomial Weighted Least Squares*

Going back to the logarithm, a further attempt was made to make it work: manually adjusting the curvature of this function to fit the data. A series of functions and manipulations were implemented to generate multiple logarithmic curves to fit the data (Figure 34); such curves were also capped at the value of 80 to further improve their results, as they would otherwise "overflow" upward. By testing each curve and deriving the best-performing one, however, it was clear the logarithm was inappropriate for the task, as the resulting function was one not resembling a logarithm at all: basically, a line always predicting the value of 80 (Figure 35). Most importantly in all of this, such function did not even perform particularly well. Thus, although even closer than before to the solution, it still did not feel just right for the task, as well as appearing forced, clumsy, and being hard to adapt altogether to different scenarios due to the logarithmic function not working with some values. In general, when too much manual input is placed in shaping the model, it is a sign of overfitting, trying to mold it excessively to perfectly fit that particular dataset rather than capturing the overall trend. Moreover, added complexity of this kind makes a model hard to reproduce in other contexts and, therefore, to adapt to various observations in the future.

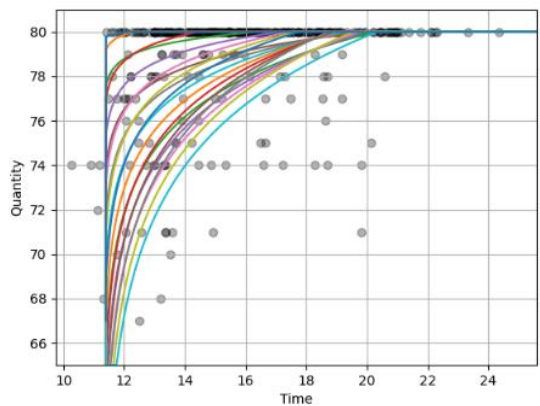

*Figure 34: Logarithm Curvature Manipulation Test*

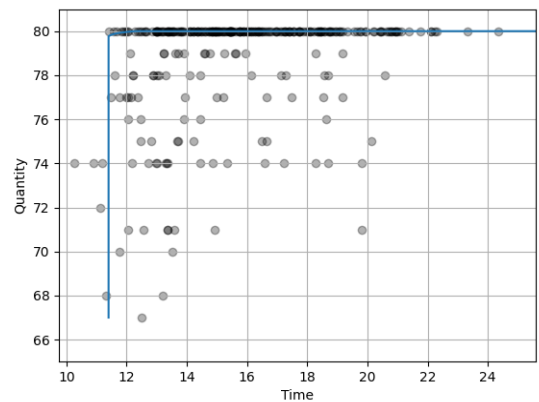

*Figure 35: Best-performing Logarithmic Curve*

The ultimate solution, in the end, came from realizing a particular feature was missing in all the functions currently used, and it was something the manual curvature tuning was actually trying to artificially replicate: an asymptote. The data naturally caps at a Y value of 80 (as no score can be higher than the provided quantity of 80 digits to memorize), stretching horizontally in that direction. Hence, the ideal function must be not only one increasing at a diminishing rate, but also one eventually leveling off at this particular value of 80.

The search, then, started for a function exhibiting this particular dual behavior, for which a candidate was eventually found in the equation of a hyperbola; after being tested (Code Excerpt 23), finally, a satisfying result was obtained.

```
def hyperbola(x, a, b):
    return a / (x + b)
popt_h, pcov_h = curve_fit(hyperbola, X_train, y_train)
y_h = hyperbola(X_test, *popt_h)
```

*Code Excerpt 23: Hyperbola-based Regression Training and Prediction*

Figures 36 and 38 report the resulting curves for the above-mentioned athletes Andrea Muzii and Alex Mullen, whilst Figures 37 and 39 show the same curves with a capping applied at the value of 80 as they still tended to slightly overflow; from now onward, these latter are the curves that will be used. Despite these curves at last looking to be the right fit, it is now time to settle these and all the other assumptions made throughout this section with model testing and validation.

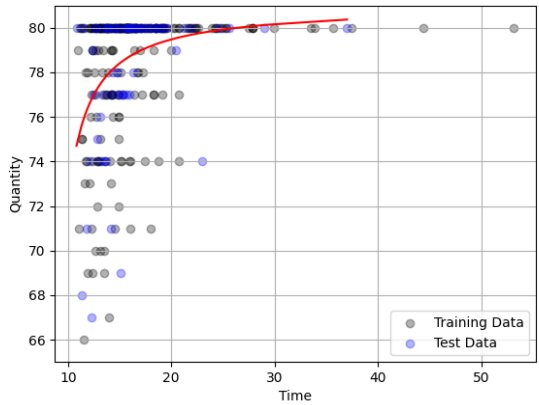

*Figure 36: Andrea Muzii's Curve*

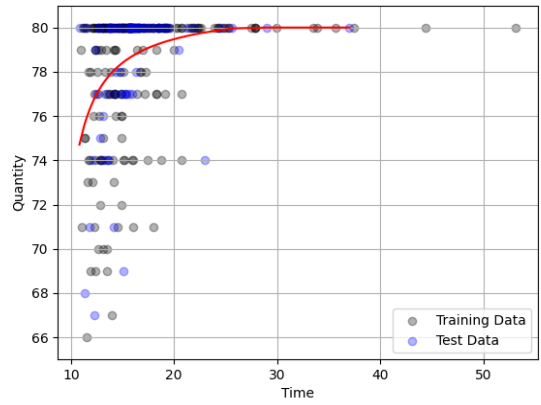

*Figure 37: Andrea Muzii's Capped Curve*

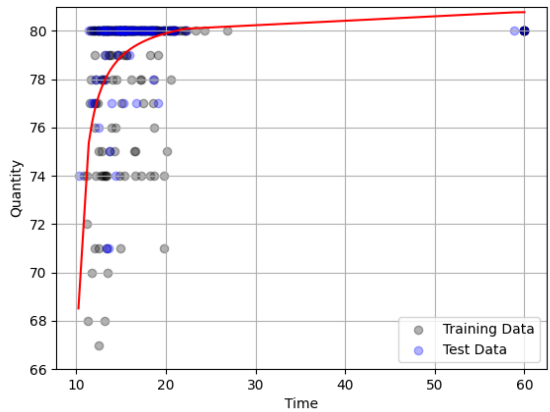

*Figure 38: Alex Mullen's Curve*

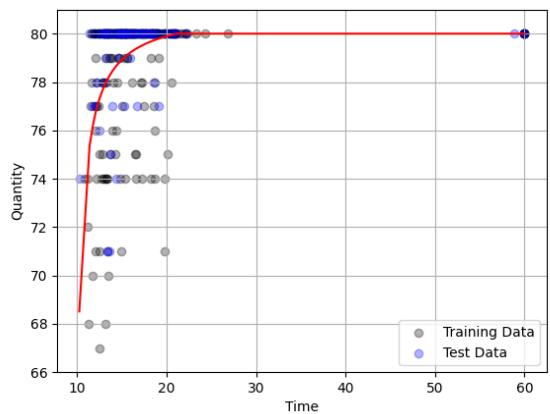

*Figure 39: Alex Mullen's Capped Curve*

## 3. Model Testing and Validation

Before comparing the models against each other, let us briefly focus on one of the best performing ones: the Decision Tree. By comparing trees with maximum depths from 1 to 10, we can see the one with a depth of 2 performs best (Figure 40). Apart from the R-squared, in fact, all these errors should be as small as possible, while the R2 itself does not even constitute a particularly valuable measure as outlined in the past chapter, because it is not appropriate for nonlinear models of this kind. Hence, this particular tree with depth equal to 2 will be taken for the overall comparison against all the other models.

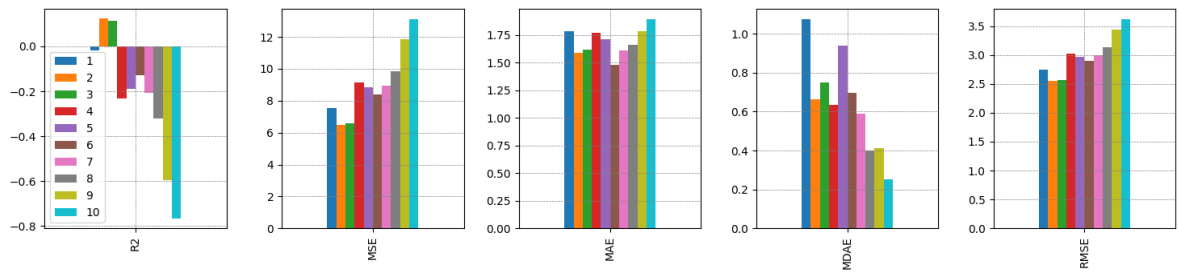

*Figure 40: Maximum Depth Comparison of Decision Trees*

```
[0]     validation_0-rmse:54.47775
[2]     validation_0-rmse:26.93614
[4]     validation_0-rmse:13.55169
[6]     validation_0-rmse:7.27029
[8]     validation_0-rmse:4.64361
[10]    validation_0-rmse:3.84485
[12]    validation_0-rmse:3.70066
[14]    validation_0-rmse:3.65461
[16]    validation_0-rmse:3.66015
[18]    validation_0-rmse:3.66939
[20]    validation_0-rmse:3.68774
[22]    validation_0-rmse:3.76680
[24]    validation_0-rmse:3.78210
```

*Code Excerpt 24: Iterations of the XGBoost Model*

A similar comparative test was made for the Random Forest models based on different numbers of estimators, yet they all performed extremely similarly. It is also interesting to show the process through which the XGBoost went (Code Excerpt 24), that is one of progressive improvement by iteratively building decision trees based on the previously fitted one.

In general, a particular trend can be outlined in the experimentation with tree-based models in this project: while the Decision Tree initially appeared promising and thus encouraged the use of a Random Forest, this latter underperformed; however, the XGBoost then slightly improved on this. Ultimately, therefore, the situation can be summarized by the plot of the MAE in Figure 41, first ascending with the forest and then partly descending again with the XGB, in terms of error magnitude.

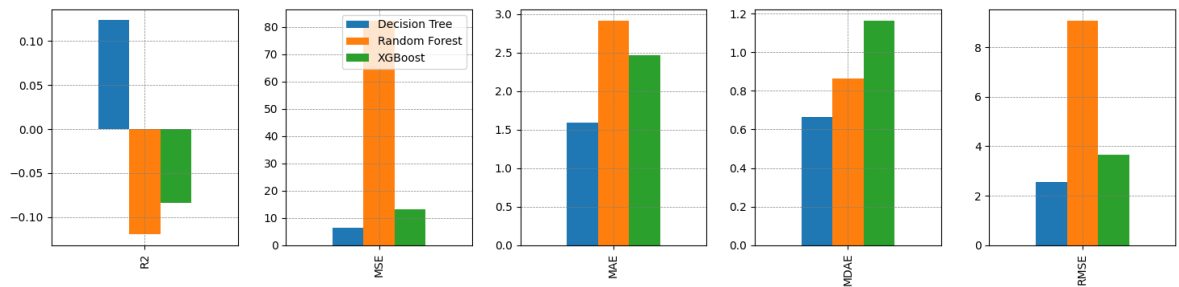

*Figure 41: Tree-based Models Comparison*

Let us now finally compare all the most significant models against each other (Table 2, Figure 42), taking the linear regression as the baseline: although the tree-based models performed rather well, it is ultimately the initial intuition of a polynomial functional curve that comes out on top. Not taking into particular consideration the R2 for the previously mentioned reasons, in fact, it is the hyperbola that achieves the lowest errors. Moreover, as mentioned earlier, this model offers the particular advantage of interpretability, which will be particularly crucial in the applications and concluding reflections of the next chapter.

| Metrics | Linear Regression | Decision Tree | Random Forest | XGBoost | Logarithm | Hyperbola |
|---|---|---|---|---|---|---|
| R2 | 0.0620264 | 0.123822 | -0.119682 | -0.084 | 0.0605606 | 0.0045648 |
| MSE | 9.45764 | 6.50741 | 82.4123 | 13.3077 | 12.7573 | 4.29184 |
| MAE | 2.32103 | 1.5892 | 2.91557 | 2.46654 | 2.60319 | 1.34789 |
| MDAE | 1.69412 | 0.662791 | 0.8635 | 1.16376 | 1.97015 | 0.89431 |
| RMSE | 3.07533 | 2.55096 | 9.07812 | 3.64798 | 3.57173 | 2.07168 |

*Table 2: Heatmap of Overall Results Comparison*

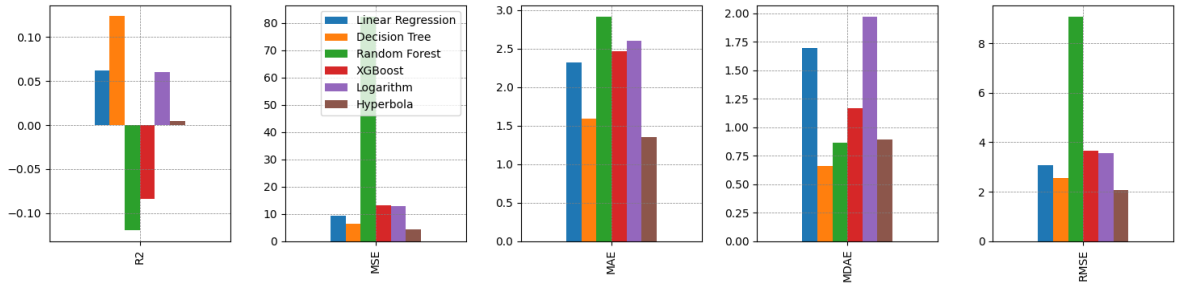

*Figure 42: Overall Results Comparison*

# CHAPTER FOUR

# RESULTS AND FUTURE DEVELOPMENTS

## 1. Results: Applications and Limitations

### 1.1. Applications

Having now actually built these software programs, let us stress again and better explain their relevance. As discussed in the first chapter, both were about finding the "perfect balance" between speed and accuracy during a memorization discipline, as the competitive context makes one have to juggle going as fast and being as precise as possible at the same time.

Task I dealt with this in relation to the IAM 5-minute Numbers discipline, for which a plane in 3D space was built predicting the universal ternary relationship between: an attempt's score, its quantity of data correctly memorized, and the highest quantity at which one would make no mistake, given the first two features. Through such function, therefore, an athlete can now input their last attempt's score and number of correctly memorized digits and obtain in return the quantity to aim at in their next competition, so as to simultaneously maximize speed and accuracy. As an illustration, Code Excerpt 25 presents some pairs of inputs from real-life attempts, together with their corresponding output from the program; to the trained eye, these relations seem reasonable, taking into account the harshness of the IAM

| [INPUTS]: | Score | Quantity | → | [OUTPUT]: | Aim |
|---|---|---|---|---|---|
| | 120 | 196 | | | 176 |
| | 140 | 178 | | | 167 |
| | 360 | 431 | | | 392 |
| | 140 | 233 | | | 207 |
| | 288 | 461 | | | 401 |
| | 160 | 476 | | | 387 |
| | 280 | 299 | | | 281 |

*Code Excerpt 25: Task I's Program Testing*

scoring system and the overall human capabilities of the athletes. Thus, the program can be deemed a success.

Task II aimed to achieve a similar objective, yet for the Memory League Numbers discipline. Moreover, rather than being universal, it performs such prediction for specific athletes, yielding their personal performance curves. With a curve of this kind, then, one is able to gauge the speed at which they start to lose accuracy or generally predict the score they would achieve at specific speeds. This, of course, can be useful to the athlete when planning for a competition or when examining their overall level at the discipline; additionally, however, it can be particularly valuable to viewers or commentators of a live match to attempt predicting the results while the athlete is still recalling the information from their memory and writing it down. After the recall phase is over, then, it can still be interesting to compare the outcome with its prediction, assessing whether it was above or below average. Such a comparison can also help spot cheaters, although rare in the sport, as an extreme deviation from their usual performance would be immediately recognizable from the graph.

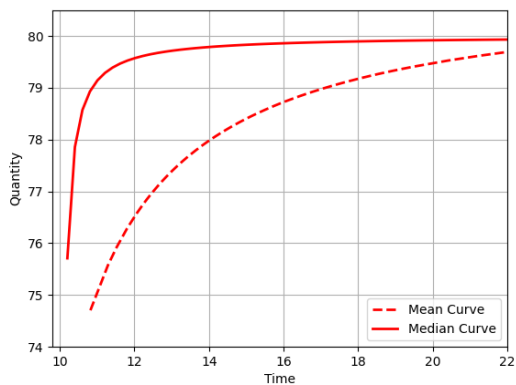
*Figure 43: Andrea Muzii's Mean vs Median Curve*

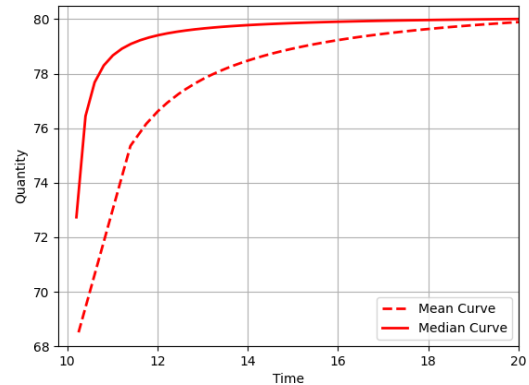
*Figure 44: Alex Mullen's Mean vs Median Curve*

Before discussing the results further, let us document an adjustment that was made to these latter. When looking at the curves obtained in the previous chapter, in fact, predictions appeared to be slightly pessimistic, indicating a level lower than the real one. In other words, they seemed to drop too early for both the athletes examined. A concept from statistics then helped settle the doubts: while the algorithm was predicting the average quantity memorized by the athlete for that given time, it was actually the median we were after. Rather than the expected value, in fact, our analysis aimed to find the most probable one. For instance, imagine an athlete achieving the following scores in five different attempts taking the exact same time: 77, 80, 80, 80, 80. Whilst the average of these quantities can be rounded to 79, the most likely outcome for that time, given these observations, is actually an 80 (which was in fact achieved in 4/5 of those attempts). To resolve the issue, then, the hyperbola-based regression model was trained again by changing its default loss function from the Mean Squared Error (MSE) to the Median Absolute Error (MAE). The results of this change are shown in figures 43 and 44.

Now, going back to the use cases of the model developed in Task II, the analysis of one's performance can also be carried out over time, to examine one's progress after several training sessions. Figure 45 illustrates this application of the software by comparing the author's performances over time. As can be seen, after a vigorous kickoff in 2021, leading to quick improvements in just a few months, the athlete then settled down to a more laid-back level as he started dedicating more to other endeavors (including this research itself) and less to the training per se.

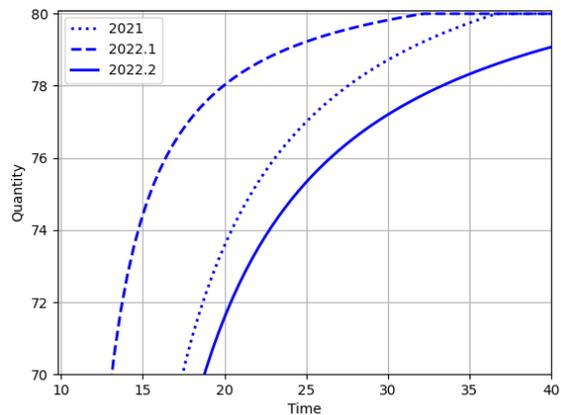

*Figure 45: Emanuele Regnani's Performance Level Over Time*

A particularly useful scope of the program developed in Task II, however, is the possibility to compare the resulting curves for multiple subjects, so as to analyze one athlete's performance against another. Perhaps, no better illustration of this could there be than the everlasting fight between the already mentioned champions Andrea Muzii and Alex Mullen in the Numbers discipline. By plotting their curves on a single graph, we can potentially settle once and for all which of the two truly is "the best" (Figure 46). The results, however, are not univocal: while Mullen appears to be just slightly better off than Muzii under the 25 seconds mark, the opposite becomes true towards the low tens. Now, although the lack of a definitive answer might feel disappointing, this particular pattern tells us something else entirely: maybe, it is this very unpredictability that makes matches between these two athletes so entertaining to watch; it is as if their levels perfectly balance each other so that, most of the time, it is ultimately a matter of pure chance whether one wins or the other.

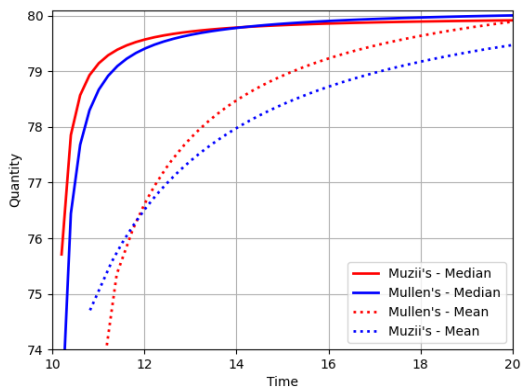

*Figure 46: Comparison Between Predicted Curves*

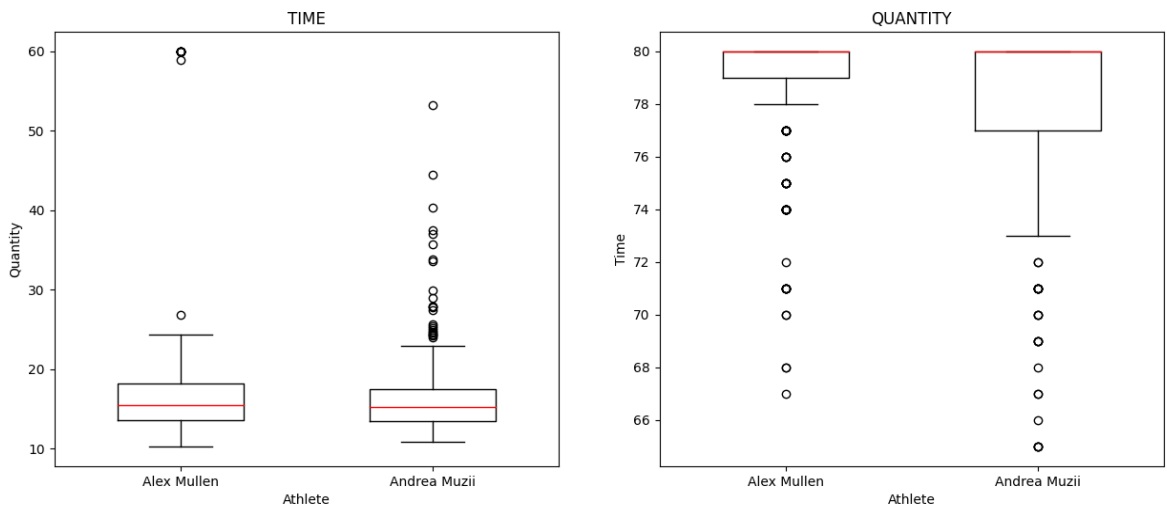

*Figure 4713: Boxplots Comparing Andrea Muzii and Alex Mullen's Data Distributions*

Similar considerations are highlighted by comparing the distributions in the athletes' data through boxplots (Figure 47): while the medians (red lines) lie extremely close to each other, the overall ranges and the outliers in each dataset differ significantly, explaining the difference between the two trends in performance.

Let us now show some specific predictions made by the algorithm, similarly to what was done above with the first software program:

```
Time [INPUT]:    →    Quantity [OUTPUT]:    Muzii's              Mullen's
                                            Mean    Median       Mean    Median
10.5                                        74      78           71      77
11                                          75      79           74      79
12                                          77      80           77      79
12.5                                        77      80           77      80
14.0                                        78      80           78      80
17.0                                        79      80           79      80
20.0                                        79      80           80      80
23.0                                        80      80           80      80
```

*Code Excerpt 26: Task II's Program Testing*

An observation could explain the main difference between the median and the mean for the athletes: as briefly touched upon in the previous chapter, for athletes memorizing digits in chunks of three at a time, an 80 might still be more likely than a 79 or a 78 when one of these latter is the predicted value because it is much rarer to forget a single or two single digits rather than an entire chunk (which would instead result in a 77). Nonetheless, these results appear to make general sense to the trained eye, hence also this second program can be considered a success overall. Yet, some remarks ought to be made, particularly regarding its limitations.

## 1.2. Limitations

In general, the software developed in Task I is rather proficient and ready to use. Conversely, the one from Task II, despite its applications demonstrated above, is more of a proof of concept than a complete product, due to its imprecisions and limitations that are to be described now.

Firstly, being solely collected from publicly rated matches, this data might not properly capture the behavior at the extreme end of the spectrum; that is, the neighborhood of the 10 seconds, in this case: athletes, in fact, tend to only truly test their limits in training sessions, as a safer performance is usually rewarded in actual games. What would be needed, then, is data also coming from solo training sessions, so as to have more varied and substantial observations pertaining to the same period. Moreover, in order to have a sufficient amount of observations to work with, data was collected from up to six years ago: this means the program might suffer from a slight "nostalgia" bias, let us say, as it might be excessively influenced by past observations when making its predictions. The solution is again the one outlined just above, as that way there would be no need to draw from exaggeratedly old data. These same shortcomings, then, make the program only suitable for experienced athletes with many public rated matches and an overall stable level of performance. Unstable ones (usually because improving rather quickly), in fact, present inconsistent data inappropriate for analysis, whilst novices or players rarely appearing in public games do not offer enough observations for the analysis to work properly no matter their actual rate of progress. Most of these limitations, however, can again be solved through a more thorough data collection process tailored to the specific athlete being analyzed.

Lastly, regarding the usability of the software itself, it can currently be accessed by the creator only, as well as being rather cluttered and unstructured. For this reason, the solution of a graphical user interface for usage by someone non-technical will be proposed shortly.

As a final remark, although not technically a true limitation of the program, the results of these algorithms are all just numerical; thus, they give no indication on how to apply what they indicate in the practice of one's mental performance. It is then the athlete who first has to train themselves and learn to control their speed and accuracy to a level at which they are able to actually follow numerical indications of this kind in the first place; alternatively, a coach figure can help with the interpretation of these results and with the resulting guidance associated to them, something perhaps an AI model could learn to do, as suggested in the second section of this chapter.

**Potential Future Upgrades**

As of its current state, the program only models the relationship between the dependent and independent variables, lacking instead a proper probabilistic model able to give the exact odds of particular outcomes happening, given the input. This could offer a better understanding of the results and exhaustive indications in general, rather than simply yielding a single number, as one would have a clearer picture of the situation and of their range of maneuver in it.

Furthermore, the software itself lacks a graphical user interface allowing anyone to utilize it, as anticipated above. Instead, it currently requires the author himself to manually tweak it to the needed use and, especially in the case of Task II, it is often missing the flexibility needed to make it work with any athlete's data, as it has so far usually been adapted to a particular athlete's scenario.

On the whole, the project met its initial objectives and can be regarded as a success. Nonetheless, the applications of the technologies utilized here to the sport deserve much greater exploration; let us then take initiative and propose some cues in this regard.

# 2. Directions for Future Research

As explained in the second chapter, the state of the art of machine learning's applications to Memory Sports has substantial room for development, as it consists of an almost completely uncharted territory. This work, then, has aimed to act as a proof of concept, a timid first stone dropped in this still lake, hoping for its ripples to stimulate progressively more advances. On this note, let us first provide a framework to think about the sector, so as to then propose several ideas for potential applications of machine learning to it.

Before even concerning about training and competition, a memory athlete has to prepare and sharpen their tools, the mnemonic devices they will use for memorization. This involves all their memory palaces, as well as the systems they adopt for numbers, cards, or even other disciplines if they so choose. Then, comes the training side of things: which methods to use and when, which psycho-physical and environmental factors matter and how, as well as analysis of previous training sessions and planning for future ones. Finally, of course, there is the competitive aspect, having to do with: optimizing one's strategy in view of either specific future opponents or general all-against-all competitions to come; planning one's goals and expectations and analyzing one's performance and results *ex post*.

These three are, perhaps, the main parts into which the sport could be divided; however, it is also characterized by a lively community and a growing public, with the resultant need to manage and engage this latter. Hence, this fourth aspect ought to also be considered for the next section.

Now, having provided a framework through which to think about this field, let us delve into the several ways machine learning could be applied to it.

## Applications of Machine Learning to Memory Sports

**Mnemonic Devices**

Starting from the mnemonic devices, machine learning could be applied to optimize one's memory palaces: an algorithm of this kind could identify what makes a location truly optimal for memorization (dimension, position, type, and so forth); it could learn to optimize the number of locations to be used in a certain room, their position with respect to one another, and the overall distribution of locations inside a sequence; it could do this whole optimization at an even grander scale, by optimizing for different disciplines, analyzing the ideal number of palaces to have for each discipline and their dimension, as well as the optimal number of days to wait for a specific palace to be usable again, based on the specific discipline it was utilized for.

Moving onto the systems used for disciplines like numbers and cards, artificial intelligence could similarly be employed to analyze each image in one's system to suggest potential improvements making it more memorable and "interoperable" with other images (as images need to interact with one another during memorizations, an aspect which was not dutifully explained here), or even if such image is viable at all for its purpose. The algorithm could also help in the initial memorization of the system by the athlete, suggesting connections to help relate the number or card to its corresponding image or even suggest an image in the first place (for instance, the King of Hearts could be represented by someone particularly successful in romance). It could also analyze the system as a whole, identifying the ideal proportion between each kind of image so as to optimize their interactions and reduce confusion (that is, how many people, animals, or objects to have). On this same note, it could spot potential duplicates (two types of cars, two similar animals, and so on) and suggest how to better distinguish them.

Of course, ideas like these would ideally involve a direct connection between one's brain, where the palaces reside and the systems are visualized, and the machine analyzing such palaces. Yet, for now, one could still at least partially tap into an idea of this kind by using services like the ones mentioned in the first chapter (particularly the visual tool that is MemoryOS) or some form of database to build such palaces and systems in a digital form for a machine to access.

Furthermore, many of these aspects would require direct real-time input from training data, so as to understand what worked and what did not in previous sessions, something that clearly complicates the program but could still potentially be viable.

**Training and Planning**

First of all in this domain, the athlete could be continuously monitored to assess, for instance, sleep quality and quantity, or hormonal and mood factors; it might seem slightly extreme, but it is already being done in physical sports and, after all, this is what is truly meant by optimizing for a sport and pushing the human limits. With all the psychophysical data collected this way, the program could adjust training sessions, or evaluate their results also based on these factors (for example, not penalize a slightly sub-average score if the biological data was particularly inadequate, and instead reward the athlete for having surpassed what was expected with such challenging mental and physical conditions, or vice versa).

The algorithm, then, could structure entire training paths tailored to the athlete and their objectives, constantly rebalancing the proportions in which each discipline is trained based on the situation, and overall prepare the practitioner for a particular competition. To illustrate, such program could balance the type of training to be done based on the period or on the objectives (while "faster than you think is possible" [28] training is generally more beneficial for improvements, one must learn to control their speed and accuracy as they approach a competition, as mistakes are usually highly penalized in such context): in a competition-free period, the software tool could suggest more ambitious quantities and speeds to aim at, so as to encourage mental adaptations and overall improvement; conversely, as a competition draws near, the program developed in Task I of this project could be employed.

Finally, it could help overcome plateaus by suggesting different methods to try or by identifying possible weak points to work on.

---

[28] *https://youtu.be/GT0wQcGU0rg?t=111*

**Competition**

Adding to what has been suggested above, machine learning could aid in achieving a particular record or beating a specific opponent, by attempting to identify and exploit the weak points of such rival. It could even perform a similar task in all-against-all competitions by analyzing the participants and their weak points and, consequently, suggesting which disciplines to focus on to attempt improving one's position in either the overall leaderboard or in discipline-specific ones.

Before a competition, then, it could provide realistic expectations of one's results given their current level; on the other hand, it could then assist the athlete in analyzing their performance *ex post*, what went wrong and what to improve for the next time.

Lastly, going back to head-to-head competitions, it could suggest which discipline to choose based on one's current psychophysical state and, perhaps, even on the state of the opponent, based on what can be seen from their results and camera feed (as well as from the opponent's own psychophysical data, if available).

**Community**

The Memory Sports community, currently scattered throughout various social media and forums, could be united under a single platform similar to what is now the Art of Memory forum[29], but with enhanced capabilities: on the forum and discussion side of things, an AI could help organize the content, assign tags to the different posts, and filter and search. It could also be employed to manage a section highlighting news in the field, new methods and techniques invented or proposed, records achieved, and anything one might want to be kept up to date with (therefore, personalizing its services to the specific user in question).

Such platform could also act as a social media for Memory Sports, suggesting which users to connect to based on similarity measures, and allowing for groups and sub-communities to emerge. It could also integrate the "competition" feature that the Memory League website allows for, all in one single platform where users could both chat together and challenge each other. Additionally, various types of competitions could be allowed for, such as team-based ones, but that is outside the scope of this text.

Next, a chatbot could be implemented to help athletes navigate the content of the forum, let them ask for advice, and to suggest ideas and solutions to issues, as well as for the public to have live commentary or receive immediate updates whenever they lose track of the score or even upon this latter changing at all in the first place. The bot could also provide interesting insights regarding the match in progress by spotting particular patterns or by offering to explain whatever the spectator might not understand about the underlying mnemonic techniques and scoring rules being used, acting as a sort of interactive "Frequently Asked Questions" section. Additionally, coupled with the newest image generation AI tools akin to DALL-E[30] and Midjourney[31], such bot could also provide visual cues together with its explanations, which would prove particularly helpful in teaching about the predominantly imaginative component of memory techniques.

---

[29] *https://forum.artofmemory.com/*
[30] *https://openai.com/product/dall-e-2*
[31] *https://www.midjourney.com/home/?callbackUrl=%2Fapp%2F*

One last interesting feature AI could be employed for is again in entertaining the public and was inspired by a comment of the two-times World Champion Johannes Mallow in an interview[32]: by having the athletes provide their systems and palaces in advance (something especially doable for the systems, which stay fixed over time), the AI could recreate for the public what is happening in the athlete's head (again, using the abovementioned tools), by showing the image corresponding to the piece of data they are currently viewing on their screen and even making it interact with both the corresponding location in the memory palace and the next image coming up.

**ChatGPT**

A special section in all of this, of course, ought to be dedicated to the revolutionary tool that is ChatGPT or other chatbots akin to it in general. In fact, some of the ideas presented hereafter were even suggested by the bot itself, when prompted about it!

A program of this nature could of course act as the previously mentioned chatbot, both for the public and for the athletes themselves, but it could even take on the role of a complete coach, managing all the steps and procedures mentioned above while "thoughtfully" guiding the athlete through the various processes and training sessions.

Pertaining to this, the bot could also provide emotional and motivational support, offering words of encouragement, reminding them of their goals, and advising them on how to overcome particular challenges. Specifically, it could aid in managing the stress and loss of focus often caused by the competitive environment.

Finally, a particular function the bot could have that has not yet been mentioned is in data generation: apart from the trivial generation of random digits, words, or playing cards, such a tool could help in generating the specific types of words needed for competitions (e.g., verbs in the infinitive, or a certain proportion of concrete nouns versus abstract ones), as well as translating them in different languages, which are all tasks that have up to now taken considerable time and money. The same goes for generating national or international names for the respective disciplines. ChatGPT can also help brainstorm new ideas for the "surprise events" popular on the Memory League platform; in this regard, AI tools in general could be used to generate completely new images and human faces to be used in competitions (and there are already examples of this being done), eliminating the barriers holding the current training software programs from extending their databases.

---

[32] *https://www.youtube.com/watch?v=jqrDICvv7wA*

# 3. Conclusion

This work has aimed to demonstrate how large the potential is for applications of machine learning to Memory Sports and mental training in general. Having provided an overview of the sport and of the machine learning to be used, a practical implementation project has been realized and explained, showing two possible use cases of this technology in the field. This final chapter, then, has sought to explain the results obtained, stress the relevance of the programs built, and highlight their limitations. Lastly, several ideas other than the two applied here have been proposed to guide future research.

As anticipated in the introduction, apart from demonstrating a possible combination of them, the ultimate intention of this study has been to fundamentally educate about the two realms of mental training and machine learning, raising genuine curiosity and interest. The latter, in fact, is among the trends and tools most powerfully and abruptly shaping today's world, hence it would be unwise to overlook it. The former, instead, could perhaps be a crucial ingredient in humans being able to cognitively stay competitive against AI, as well as the basis of the critical skill that is learning in the so rapidly-changing environment we live in. Moreover, mental training in general could be the tool through which we maintain our authority and control over machines rather than the other way around, as striving to keep our cognitive faculties sharp would result in minds considerably more conscientious and thus harder to subject.

Combining these two realms, then, would result in perhaps the ultimate weapon against artificial intelligence taking our jobs, as we would then be able to couple the numerical abilities of the machine with the emotional, social, and creative capabilities of the human. Ultimately, in fact, by not only engaging in mental training but by also enhancing it through the power of AI, we could reach that human-machine symbiosis that the two entities on their own simply cannot rival and which is anyway the future we are moving towards as a species.

# BIBLIOGRAPHY

- Baldwin R.; 2019: *"The Globotics Upheaval"*

- Brown P., Roediger H., McDaniel M.; 2014: *"Make it Stick: The Science of Successful Learning"*

- De Concini A., Muzii A., De Luca V.; 2022: *"Mnemonica 2.0"*
  *https://www.alessandrodeconcini.com/tutti-i-corsi/corso-tecniche-memoria-mnemonica*

- De Luca V.; 2017: *"Una mente prodigiosa"*

- Domingos P.; 2015: *"The Master Algorithm"*

- Géron A.; 2019: *"Hands-on Machine Learning with Scikit-Learn, Keras & TensorFlow"* (2nd edition)

- Harari Y. N.; 2011: *"Sapiens"*

- Newport C.; 2016: *"Deep Work"*

- Wilson J., Daugherty, P. R.: *"Collaborative Intelligence: Humans and AI Are Joining Forces"* - Harvard Business Review
  *https://hbr.org/2018/07/collaborative-intelligence-humans-and-ai-are-joining-forces*

- Young S. H.; 2019: *"Ultralearning: Master Hard Skills, Outsmart the Competition, and Accelerate Your Career"*

**Secondary Sources**

- Banko M., Brill E.; 2001: *"Scaling to very very large corpora for natural language disambiguation"*
  *https://dl.acm.org/doi/10.3115/1073012.1073017*

- Frost J.: *"R-squared Is Not Valid for Nonlinear Regression"*
  *https://statisticsbyjim.com/regression/r-squared-invalid-nonlinear-regression/*

- Frost J.: *"The Difference between Linear and Nonlinear Regression Models"*
  *https://statisticsbyjim.com/regression/difference-between-linear-nonlinear-regression-models/*

- Kaul S.: *"Model Selection Techniques in ML/AI with Python"*
  *https://medium.com/analytics-vidhya/model-selection-techniques-in-ml-ai-with-python-fdf308d9fa10*

- Muzii A.; 2022: *"Come funzionano le gare di memoria"*
  *https://www.youtube.com/watch?v=k4t2cgQOwHU*

- Scikit-learn Python Library
  *https://scikit-learn.org/stable/*

- Shukla P.: *"A Comprehensive Guide to Logarithmic Regresion"*
  *https://heartbeat.comet.ml/a-comprehensive-guide-to-logarithmic-regression-d619b202fc8*

- Wikipedia - Mnemosyne
  *https://en.wikipedia.org/wiki/Mnemosyne*

- Wikipedia - Polynomial Regression
  *https://en.wikipedia.org/wiki/Polynomial_regression*

- Wikipedia - The Nine Muses
  *https://en.wikipedia.org/wiki/Muses*

**Main Memory-related Websites Mentioned**

- Canadian Mind Sports Association
  *https://canadianmindsports.com/memory-sports/*
- International Association of Memory (IAM)
  *https://www.iam-memory.org/*
- Memory League
  *https://memoryleague.com/#!/home*
- Official Rulebook of the International Association of Memory (IAM)
  *https://www.iam-memory.org/wp-content/uploads/2019/11/IAM_Rulebook_Nov2019.pdf*
- USA Memory Championship
  *https://www.usamemorychampionship.com/*

**Briefly Mentioned Works**

- Aeschylus; 424 BCE: *"Prometheus Bound"*
- Alighieri D.; 1320: *"The Divine Comedy"*
- Juvenal; 100-127 AD: *"Satire X"*
- Plato; 370 BCE: *"Phaedrus"*
- Svevo I.; 1923: *"Zeno's Conscience"*

# LIST OF FIGURES

# LIST OF TABLES

# LIST OF CODE EXCERPTS